\documentclass[conference]{IEEEtran}
\IEEEoverridecommandlockouts
\usepackage{cite}
\usepackage{amsmath,amssymb,amsfonts}
\usepackage{algorithmic}
\usepackage{graphicx}
\usepackage{textcomp}
\usepackage{xcolor}
\usepackage{float}
\def\BibTeX{{\rm B\kern-.05em{\sc i\kern-.025em b}\kern-.08em
    T\kern-.1667em\lower.7ex\hbox{E}\kern-.125emX}}
\usepackage{amsthm}
\usepackage{booktabs,subcaption,amsfonts,dcolumn}
\usepackage{algorithm}
\usepackage{algorithmic}
\usepackage{stfloats}

\theoremstyle{definition}
\newtheorem{example}{Example}[section]

\usepackage{fancyhdr}
\usepackage{url}
\usepackage{hyperref}

\begin{document}

\title{Class-Specific Attention (CSA) for Time-Series Classification}

\author{\IEEEauthorblockN{Yifan Hao}
\IEEEauthorblockA{\textit{Department of Computer Science} \\
\textit{New Mexico State University}\\
Las Cruces, NM, US \\
yifan@nmsu.edu}
\\\vspace{0.18in}
\IEEEauthorblockN{Jiefei Liu}
\IEEEauthorblockA{\textit{Department of Computer Science} \\
\textit{New Mexico State University}\\
Las Cruces, NM, US \\
jiefei@nmsu.edu}
\and
\IEEEauthorblockN{Huiping Cao}
\IEEEauthorblockA{\textit{Department of Computer Science} \\
\textit{New Mexico State University}\\
Las Cruces, NM, US \\
hcao@nmsu.edu}
\\\vspace{0.18in}
\IEEEauthorblockN{Huiying Chen}
\IEEEauthorblockA{\textit{Department of Computer Science} \\
\textit{New Mexico State University}\\
Las Cruces, NM, US \\
hchen@nmsu.edu}
\and
\IEEEauthorblockN{K. Sel\c{c}uk Candan}
\IEEEauthorblockA{\textit{School of Computing, Informatics,} \\
\textit{and Decision Systems Engineering} \\
\textit{Arizona State University}\\
Tempe, AZ, US \\
candan@asu.edu}
\\
\IEEEauthorblockN{Ziwei Ma}
\IEEEauthorblockA{\textit{Department of Mathematics} \\
\textit{University of Tennessee at Chattanooga}\\
Chattanooga, TN, US \\
ziwei-ma@utc.edu}
}

\maketitle


\begin{abstract}
Most neural network-based classifiers extract features using several hidden layers and make predictions at the output layer by utilizing these extracted features. We observe that not all features are equally pronounced in all classes; we call such features {\em class-specific features}. Existing models do not fully utilize the class-specific differences in features as they feed all extracted features from the hidden layers equally to the output layers. Recent attention mechanisms allow giving different emphasis (or attention) to different features, but these attention models are themselves class-agnostic. In this paper, we propose a novel class-specific attention (CSA) module to capture significant class-specific features and improve the overall classification performance of time series. The CSA module is designed in a way such that it can be adopted in  existing neural network (NN) based models to conduct time series classification. In the experiments, this module is plugged into five start-of-the-art neural network models for time series classification to test its effectiveness by using 40 different real datasets. Extensive experiments show that an NN model embedded with the CSA module can improve the base model in most cases and the accuracy improvement can be up to 42\%. 
Our statistical analysis show that the performance of an NN model embedding the CSA module is  better than the base NN model on 67\% of MTS and 80\% of UTS test cases and is significantly better on 11\% of MTS and 13\% of UTS test cases.
\end{abstract}


\section{Introduction}
Time series analysis (e.g., classification, anomaly detection) is becoming more critical in applications that collect increasing amount of time series data.
%
Different neural network (NN)-based approaches, such as  Long Short-Term Memory (LSTM) model~\cite{ref:lstm_neco1997}, Convolutional Neural Networks (CNNs)~\cite{ref:dcnn_ijcai2015}, have been introduced for time series analysis. Fully-Convolutional Networks (FCN) are considered as a strong baseline in time series classification~\cite{ref:fcn_ijcnn2017}.
Most neural network models, either CNN or LSTM based, have several {\em hidden layers} (e.g., convolutional layers or LSTM layers) to generate temporal features -- only the last several layers (e.g., global pooling layers or fully-connected layers) are used as {\em output layers} to make predictions~\cite{ref:cross_attn_ijcai2020,ref:lstm_fcn_acc2018,ref:mlstm_fcn_nn2019,ref:dcnn_ijcai2015,ref:mccnn_fcs2016}. 

Naturally, each extracted feature may not equally contribute to the time series  classification of instances to different classes~\cite{ref:psv_tkde2019}: some features may be more effective for one class while other features may be more pronounced in instances of another class. If one feature mostly comes from instances in one specific class, in this paper, we refer to such feature as a {\em class-specific feature}.
 \begin{figure}[htb]
 \begin{center}
 \includegraphics[width=0.55\linewidth, 
 ]{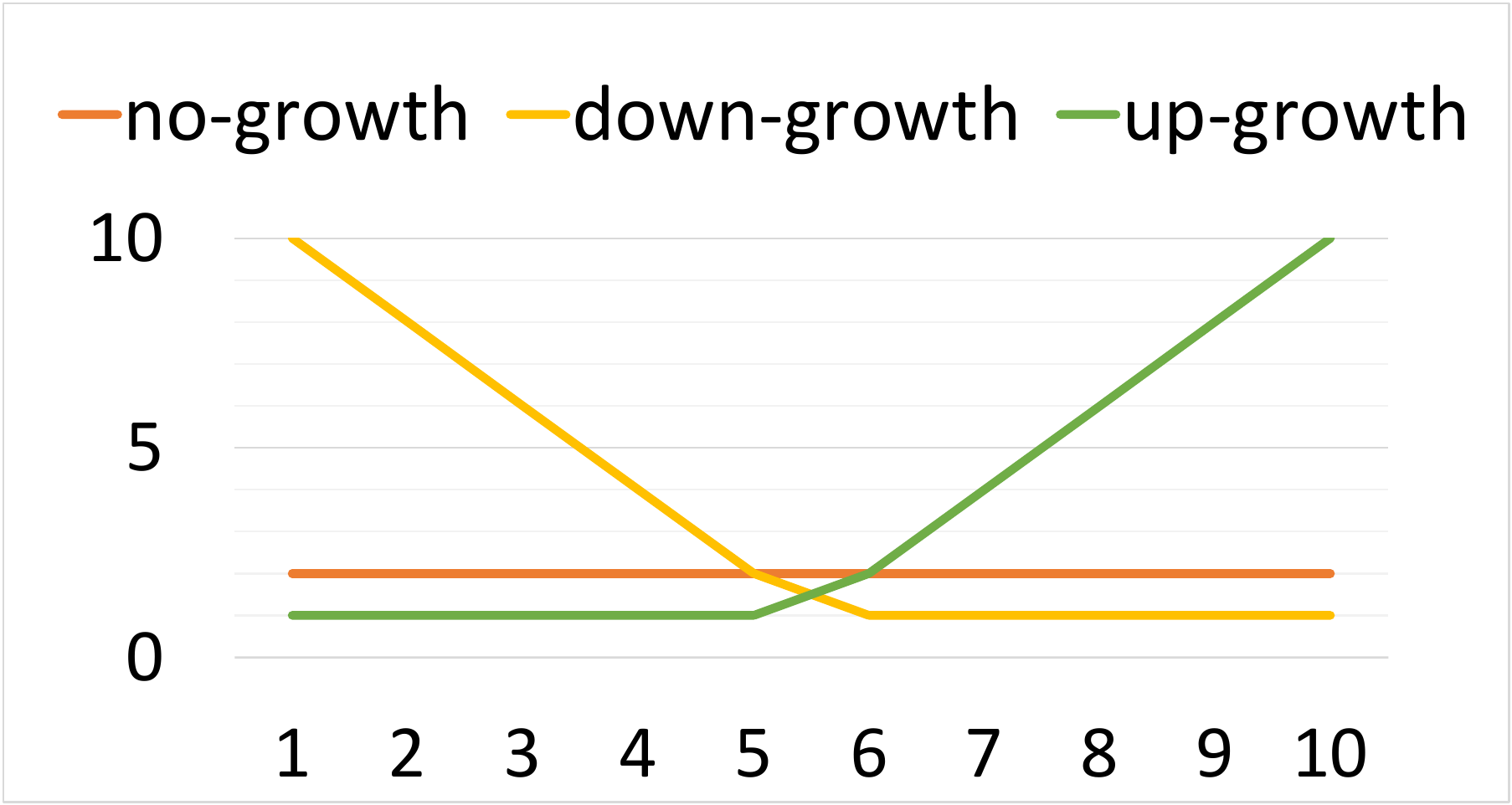}
 \caption{Toy time-series examples from different classes}
 \label{fig:ts_exp}
 \end{center}
 \end{figure}

\begin{example}[Class-specific features]
\label{exp:ts_exp}
Fig.~\ref{fig:ts_exp} shows the stock prices of three  companies. The red curve shows the stock prices from a stable performing but no-growth 
company, without any price changes. 
The green curve shows the prices from an up-growth company and the yellow curve shows the prices of a down-growth company. 
It is easy to see that, in this example, the sub-sequences from 
time step 1 to 5 can easily separate the down-growth company from the other two; sub-sequences from 6 to 10 would be more helpful to differentiate 
the up-growth company from
the other two companies; and the whole time range from 1 to 10 may be needed to differentiate the no-growth company from the down-growth and up-growth companies.
\end{example}

While the existence of class-specific features have been acknowledged and leveraged in the context of improving model explainability~\cite{ref:psv_tkde2019}, unfortunately they are not well-utilized for the primary classification tasks by the 
neural networks models, which generally feed all the features generated by the hidden layers to the output layers~\cite{ref:lstm_fcn_acc2018,ref:mlstm_fcn_nn2019,ref:fcn_ijcnn2017}. 
TapNet~\cite{ref:tapnet_aaai2020} is the most recent model that utilizes class-label information to classify time series.


In recent years, attention mechanisms~\cite{ref:first_attn_iclr2015,ref:cross_attn_ijcai2020,ref:google_all_atte_nips2017} have been introduced and widely leveraged to allow providing different levels of emphasis (or attention) to different features. Extensive research has shown that such attention mechanisms can improve the performance of neural network models in different applications. However, these attention mechanisms themselves tend to be class-agnostic and, to the best of our knowledge, for a general time-series classification task, no existing attention mechanism explicitly incorporates class-specific features for generating emphasis.


To make use of the class-specific features to improve the performance of general NN models to classify time series data, this paper proposes a novel Class-Specific Attention (CSA) module that captures class-specific features to learn class-specific attention, without negatively impacting the testing phase. 
The CSA is designed to give emphasis (using attention) to different class-specific features and   differentiates the features that are particularly important to specific classes from the features that commonly occur in instances for all the classes.
The CSA module is different from existing attention designs~\cite{ref:first_attn_iclr2015,ref:google_all_atte_nips2017,ref:cross_attn_ijcai2020} because CSA utilizes class labels to calculate attention. 

The main {\bf contributions} of this paper are: 
\begin{itemize}
\item
We propose a Class-Specific Attention (CSA) module to capture class-specific features and learn class-specific attention from (multivariate) time series data.  This module is general and can be adopted in most neural network based models that are designed for time series classification 
(with or without attention) to improve their performance.
\item 
We design a Class-differentiation (CD) component to better identify the specific and important features to one class. The CSA module learns class-specific features in the hidden layers during training and can be easily used
in testing.
\item 
The CSA module is applied to five state-of-the-art neural network models on 40 real datasets (28 multivariate time-series (MTS) and 12 univariate time-series (UTS)). Extensive experiments show that a model with CSA module provides better overall performance (with accuracy improvement up to $42\%$) compared to the same model without CSA. Statistical analysis shows that the CSA module improves a base model on 67\% of MTS tests
and 80\% of the UTS tests, is significantly better than a base
base on 11\% of MTS and 13\% of UTS tests.
\end{itemize}

The rest of this paper is organized as follows. Section~\ref{sec:rel} discusses the related works. Section~\ref{sec:pro} 
presents the architecture, the learning algorithm, and the utilization of the newly proposed CSA module.
Section~\ref{sec:exp} 
reports our experimental results to evaluate the performance of the CSA module.
Section~\ref{sec:con} concludes this work and 
discusses future directions.

\section{Related Works}
\label{sec:rel}
Time series classification is a well-studied problem with extensive literature on the topic. 
Recently, deep learning models 
have shown great success in time series classification. 
Convolutional Neural Networks (CNNs) 
have been exploited in time series classification 
because shallow
convolutional 
layers can capture short-term temporal dependencies and deep 
convolutional
layers can 
encode long-term temporal dependencies. 
Various CNN models are designed to classify multivariate time series data~\cite{ref:dcnn_ijcai2015,ref:mccnn_fcs2016}. In~\cite{ref:fcn_ijcnn2017}, a Fully-Convolutional Network (FCN) is proposed. 
It is considered as a strong baseline for time series classification. Different from traditional CNN models, FCN contains a group of convolutional layers without any pooling operations to prevent overfitting. Features from the convolutional layers are fed into a global averaged pooling layer to make predictions.

Other than neural networks using convolutional operations, the Long Short-Term Memory (LSTM)~\cite{ref:lstm_neco1997,ref:drnn_arxiv2014, ref:dual_attn_rnn_ijcai2017} and Gated Recurrent Unit (GRU)~\cite{ref:gru_2014} based models are also widely
used in temporal analysis. LSTM is one specific type of Recurrent Neural Network (RNN) models and GRU is a simpler and faster version of LSTM with similar performance. 
Both LSTM and GRU capture 
temporal features and makes predictions by using gating functions in their dynamics states.

%
To make good use of the FCN and LSTM models, Karim et al.~\cite{ref:lstm_fcn_acc2018,ref:mlstm_fcn_nn2019} further propose two state-of-the-art models, LSTM-FCN and MLSTM-FCN,  by combining both FCN and LSTM to classify uni-variate and multivariate time series. 


Recently, the attention mechanism~\cite{ref:first_attn_iclr2015} has been widely 
deployed on neural network classifiers to solve various types of problems. 
The design of the attention mechanism is motivated by human visual attention. For example, when looking at an image or reading a sentence, a human can focus on some regions (with ``high resolution'') in an image or some keywords in a sentence. Similarly, utilizing attention design in neural networks can help the model to focus on more important features and pay less attention to other features.
Attention is first designed in image recognition domain because it is motivated by how a human pays visual attention to different regions of an image. 
In recent years, 
attention mechanisms have been applied to language translation~\cite{ref:google_all_atte_nips2017}, speech recognition~\cite{ref:attn_voice_2017}, and time series analysis~\cite{ref:cross_attn_ijcai2020,ref:lstm_fcn_acc2018,ref:mlstm_fcn_nn2019,ref:dual_attn_rnn_ijcai2017}. 
In attention mechanisms, 
the features of input data are kept in a hidden key space and a hidden value space (e.g., key-value pairs for a database of research articles). Given a user request (e.g., target sentence in natural language translation domain) or query (e.g., user queries in retrieval systems), the request features are represented in the hidden query space. 
Attention values are calculated using the similarity of the key features and the query features and are applied to adjust the features in the hidden value space. 
%
Given a group of features, attention can be used to evaluate the feature importance based on different queries. Despite all the success that attention mechanisms bring, existing attention mechanisms cannot be directly applied to capture class-specific features.

Using class-label information in classification models has been explored.
Some works (\cite{ref:psv_tkde2019} and \cite{ref:cnn_class_specific}) 
apply class information for well-trained models to improve either the model explainability or the classification performance. 
TapNet~\cite{ref:tapnet_aaai2020}
uses class label information to build a prototype (a component inside their model) for each class; they average the features of the instances for the same class label and apply softmax on the different prototypes. This is different from our design of calculating attention from  class-specific key and query features~\cite{ref:first_attn_iclr2015}.

\begin{figure*}[t]
\begin{center}
\includegraphics[width=0.8\linewidth,height=4cm,scale=0.2]{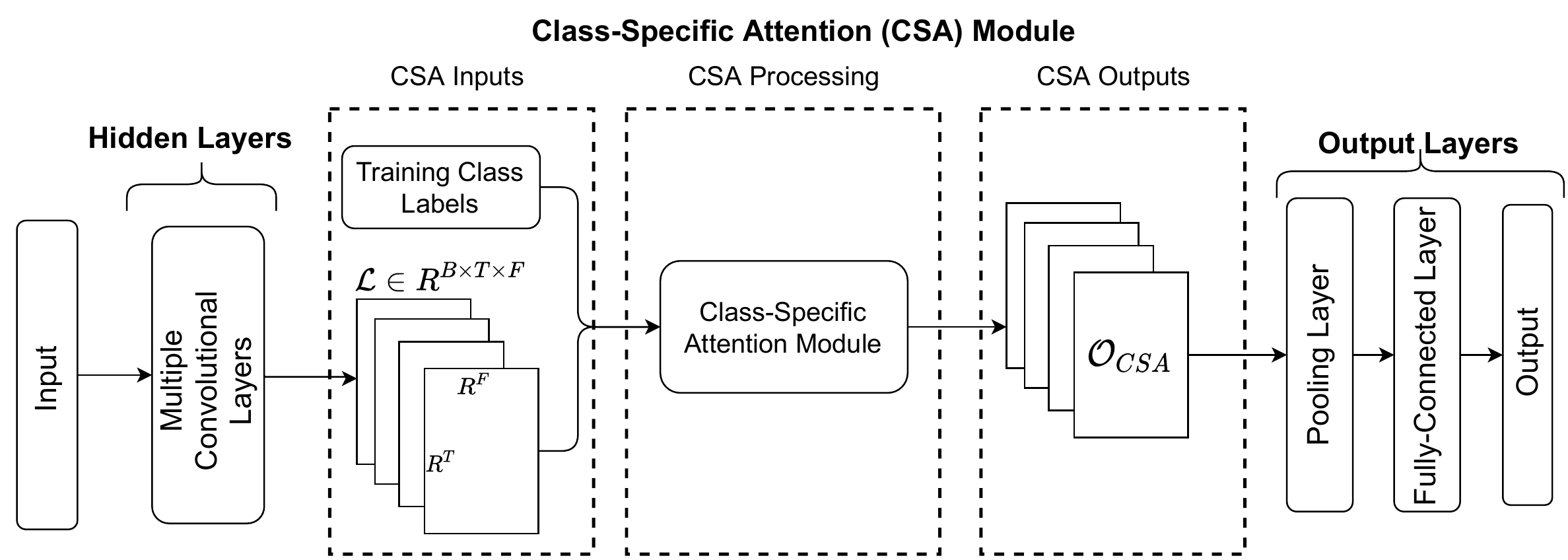}
\caption{The overview of a FCN Model, extended with Class-Specific Attention (CSA)
}
\label{fig:csa_bigpicture}
\end{center}
\end{figure*}

\section{Class-Specific Attention (CSA)}
\label{sec:pro}

%
It is possible that features generated from hidden 
layers of neural network models do not equally contribute to classifying different classes. Furthermore, features that are important to classify instances belonging to one class may not be useful in classifying instances belonging to other classes~\cite{ref:psv_tkde2019}. 
This phenomenon becomes more obvious in time-series classification when a sub-sequence of a time series, instead of the whole sequence, is more useful to classify instances from specific classes
(as in Eg.~\ref{exp:ts_exp}).
In this section, we describe a {\em Class-Specific Attention (CSA)} module that builds on the
above observations to identify and leverage class-specific features in time series classification.  

\subsection{Overall Design}
We explain the core idea and overall design of the CSA module using  Fig.~\ref{fig:csa_bigpicture}.
The CSA module is designed in a way such that it can be adopted in 
existing neural network (NN) based time series classification models.
Without loss of generality, we use the  FCN~\cite{ref:fcn_ijcnn2017} model as a representative NN model to explain the design of our newly proposed CSA module.
In this figure, the CSA module is used to extend the FCN model for time series classification for illustration purposes. The CSA module is shown in the three dotted-line boxes at the center of the figure. Note that, while CSA can be placed after any hidden layer of a neural network model,
in this figure (and in the experiments reported in the paper), we place it after the last hidden layer of the neural model, as later layers capture more higher-level features than the previous layers~\cite{ref:psv_tkde2019}. CSA takes a special form of input that leverages the class label information of training instances -- such labels are generally used by existing neural network models for evaluating the loss value from the output layer, but are not explicitly utilized for learning hidden features. In contrast to the state-of-the-art, the CSA input data integrates class label information of training instances with the features learned from previous hidden layers.


CSA gives different 
attention (or emphasis) to features generated in the hidden layers to learn class-specific features, which 
are then used to evaluate the probabilities of the data instances belonging to different classes.
The design of CSA adopts the basic idea of self-attention mechanism~\cite{ref:google_all_atte_nips2017} where the features in the three hidden spaces (key, value, query) are learned  from the same input data. 
%
%
%
%
%
On the other hand, CSA  differs from self-attention mechanisms 
in two major aspects: 
\begin{itemize}
\item First, CSA is designed to learn the class-specific features in the training stage and preserve these features in the {\em hidden query space}. 
Yet, the design of the query space, which encodes the class-specific features, can be directly utilized in the testing stage without knowing the class labels of testing instances. 
%
%
%
\item Typical attention mechanisms calculate attention values by using the similarity between the key features and the query features; however, such a calculation cannot differentiate features that commonly exist in all the instances from the class-specific features that only exist in instances of specific classes. In contrast, attention calculation in the CSA module  {\em differentiates the importance of class-specific features from other features}. 
\end{itemize}
Table~\ref{tb:symbol} lists the symbols 
used in the method description.

\begin{figure*}[htb]
\begin{center}
\includegraphics[width=0.9\linewidth]{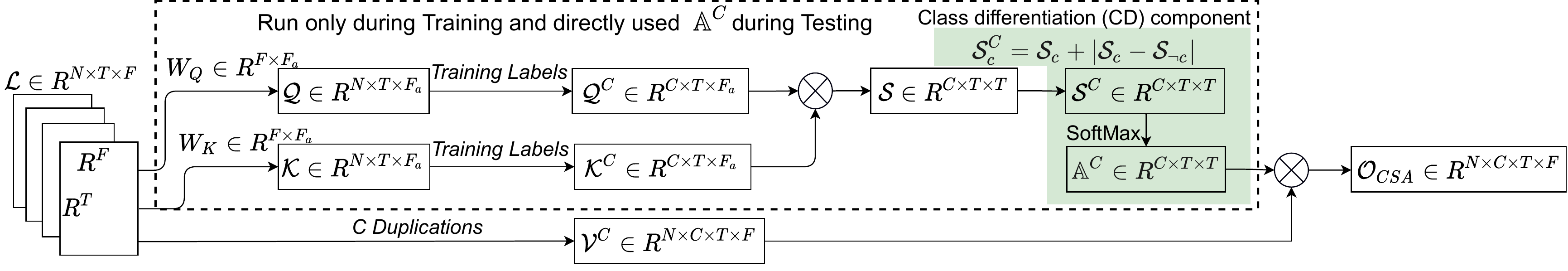}\\
$\otimes$ denotes matrix multiplication and $\oplus$ denotes matrix addition
\end{center}
\caption{The architecture of the Class-Specific Attention (CSA) Module -- notation convention: (1) A symbol using AMS blackboard bold font ($\mathbb{A}^C$) represents a tensor learned from training instances, and directly used for testing; (2) the other tensors are represented using math calligraphic font style (e.g., $\mathcal{L}$, $\mathcal{K}$, $\mathcal{V}$, $\mathcal{S}$); (3) class-specific tensors are denoted with a superscript $C$ (e.g., $\mathcal{Q}^C$, $\mathcal{K}^C$, $\mathcal{V}^C$, $\mathcal{S}^C$); (4) kernel weights are represented using $W$ with a subscript ($W_Q$, $W_K$); (5) dimension sizes are represented as superscript for $R$ (e.g., $R^{N\times T}$).
}
\label{fig:csa_details}
\end{figure*}

\begin{table}[htb]
\centering
\small
\begin{tabular}{c | l}
\toprule
Symbol & Meaning\\
\midrule
$N$ & \# of instances in a TS dataset\\
$C$ & \# of distinct classes in a dataset\\
$V$ & \# of variables in a TS dataset \\
$T$ & \# of time points of one time series in $X$\\\hline
$\mathcal{L}$ & Feature matrix from hidden layers
\\
$F$ & \# of features at each time point in $X$\\
$F_a$ & \# of features in the hidden key and query spaces\\
$B$ & \# of instances in one batch \\
\bottomrule
\end{tabular}
\vspace{0.05in}
\caption{Symbols used in this paper}
\label{tb:symbol}
\end{table}




\begin{algorithm}[htb]
\caption{{\em CSA\_Calculation}}
\label{alg:csa_cal}
\textbf{Input}: $\mathcal{L}\in R^{N\!\times\! T\!\times\!F}$: feature tensor from the hidden layers\\
\textbf{Output}: $\mathcal{O}_{CSA}\in R^{N\!\times\! C \!\times\! T\!\times F}$: feature tensor adjusted using class-specific attention
\begin{algorithmic}[1] 
\STATE \label{csa_cal:key} $\mathcal{K}=\mathcal{L} \cdot W_K$  where $W_K \in R^{F\times F_a}$ \label{alg:learnK}
\STATE \label{csa_cal:query} $\mathcal{Q}=\mathcal{L} \cdot W_Q$ where $W_Q \in R^{F\times F_a}$ \label{alg:learnQ}
\STATE Initialize $\mathcal{K}^C \in R^{C\!\times T\!\times F}, \mathcal{Q}^C \in R^{C\!\times T\!\times F}$
\FOR{each class label $c$}\label{alg:KcQc_start}
    \STATE $\mathcal{L}_c$ = all the instances belonging to class $c$. 
    \STATE \label{csa_cal:class_key}
    $\mathcal{K}^C[c,:,:]=average(\mathcal{K}[\mathcal{L}_c, :, :]$) on the first dimension
    \STATE \label{csa_cal:class_query} $\mathcal{Q}^C[c,:,:]=average(\mathcal{Q}[\mathcal{L}_c, :, :])$ on the first dimension
\ENDFOR \label{alg:KcQc_end}
\STATE $\mathcal{S}=\mathcal{K}^C \cdot (\mathcal{Q}^C)^T$ where $\mathcal{S}\in R^{C\times T\times T}$ \label{csa_cal:getS}
\STATE Initialize $\mathcal{S}^C \in R^{C\!\times T\!\times T}$
\FOR{each class label $c$}\label{csa_cal:s_diff1}
    \STATE $\mathcal{S}_c=\mathcal{S}[c, :, :]$
    \STATE $\mathcal{S}_{\neg c}=\frac{SUM(\mathcal{S})-\mathcal{S}_c}{C-1}$
    \STATE \label{csa_cal:sc_diff} $\mathcal{S}^C[c, :, :]= \mathcal{S}_c + abs(\mathcal{S}_c-\mathcal{S}_{\neg c})$
\ENDFOR \label{csa_cal:s_diff2}
\STATE \label{csa_cal:a_final} $\mathbb{A}^C=SoftMax(\mathcal{S}^C) \in R^{C\!\times T\!\times T}$ \label{csa_cal:getA}
\STATE \label{csa_cal:value} $\mathcal{V}=\mathcal{L}\cdot W_V$, where $W_V \in R^{F\times F}$
\STATE $\mathcal{V}^C=$ shape $C$ copies of $\mathcal{V}$ to $R^{N\times C\times T\times F}$
\STATE $\mathcal{L}^C=$ shape $C$ copies of $\mathcal{L}$ to $R^{N\times C\times T\times F}$
\STATE $\mathcal{O}_{CSA}=\mathcal{L}^C + \sigma\times (\mathbb{A}^C \cdot \mathcal{V}^C)$ \label{csa_cal:getOcsa}
\STATE \textbf{return} $\mathcal{O}_{CSA}$
\label{csa_cal:return}
\end{algorithmic}
\end{algorithm}

\subsection{Overview of the CSA Architecture}
\label{sec:csa_cal}

%
%
%
The features from the hidden layers of the FCN model are fed as input to the CSA module 
-- in particular, we assume that features are generated by the hidden layers that are immediately before the CSA module in the architecture.

%
%
%
%
%
%
The CSA architecture outlined in Fig.~\ref{fig:csa_details} embodies our key design decisions: 
\vspace{0.05in}

\noindent{\bf Design Decision $\#1$.} 
Unlike the existing self-attention mechanisms,  
 CSA introduces the class-specific attention $\mathbb{A}^C$ to differentiate class-specific features from other features.
%
%
%
In particular, after learning the 
features $\mathcal{Q}$ and $\mathcal{K}$ in the latent query and key spaces,  
CSA aggregates the instances in the same class in these two feature spaces to get two class-specific feature tensors $\mathcal{Q}^C$ and $\mathcal{K}^C$
so that they can be further utilized in calculating the class-specific attention $\mathbb{A}^C$ and the updated features $\mathcal{O}_{CSA}$.
%
\vspace{0.05in}

\noindent{\bf Design Decision $\#2$.} 
The class-specific attention $\mathbb{A}^C$ is designed to be a global tensor. It is updated during the training stage 
and can be directly used in testing. Consequently, the CSA design avoids the need for the  testing instances to have class labels.

\subsection{Training Process}
\label{sec:csa_details}

The ultimate goal of the CSA training  is to convert regular features (denoted as $\mathcal{L}$ in Fig.~\ref{fig:csa_details}) to  class-specific features ($\mathcal{O}_{CSA}$ in Fig.~\ref{fig:csa_details}). This conversion utilizes the class-specific attention $\mathbb{A}^C$. 

The {\em CSA\_Calculation}
algorithm, shown in Algorithm~\ref{alg:csa_cal},
provides an overview of the learning process for CSA training.
First, for the training instances, 
the features in the hidden query space ($\mathcal{Q}$) and key space ($\mathcal{K}$) are learned (Lines~\ref{alg:learnK}-\ref{alg:learnQ}).
Then, these two feature tensors are transformed to class-specific query features $\mathcal{Q}^C$ and class-specific key $\mathcal{K}^C$ features  (Lines~\ref{alg:KcQc_start}-\ref{alg:KcQc_end}). 
Next, $\mathcal{Q}^C$ are combined with $\mathcal{K}^C$ to calculate the class-specific attention $\mathbb{A}^C$ (Lines~\ref{csa_cal:getS}-\ref{csa_cal:getA}). 
Finally, the attention $\mathbb{A}^C$ is applied to the value features 
$\mathcal{V}^C$ (which is reshaped from the original feature tensor $\mathcal{L}$)
%
to get $\mathcal{O}_{CSA}$ (Line~\ref{csa_cal:getOcsa}).

In what follows, we provide details of the learning processes for calculating 
the class-specific attention $\mathbb{A}^C$. 

\subsubsection{Generating Class-Specific Features $\mathcal{Q}^C$ and $\mathcal{K}^C$}
\begin{figure}[htbp]
\begin{center}
\includegraphics[width=1\linewidth,scale=1]{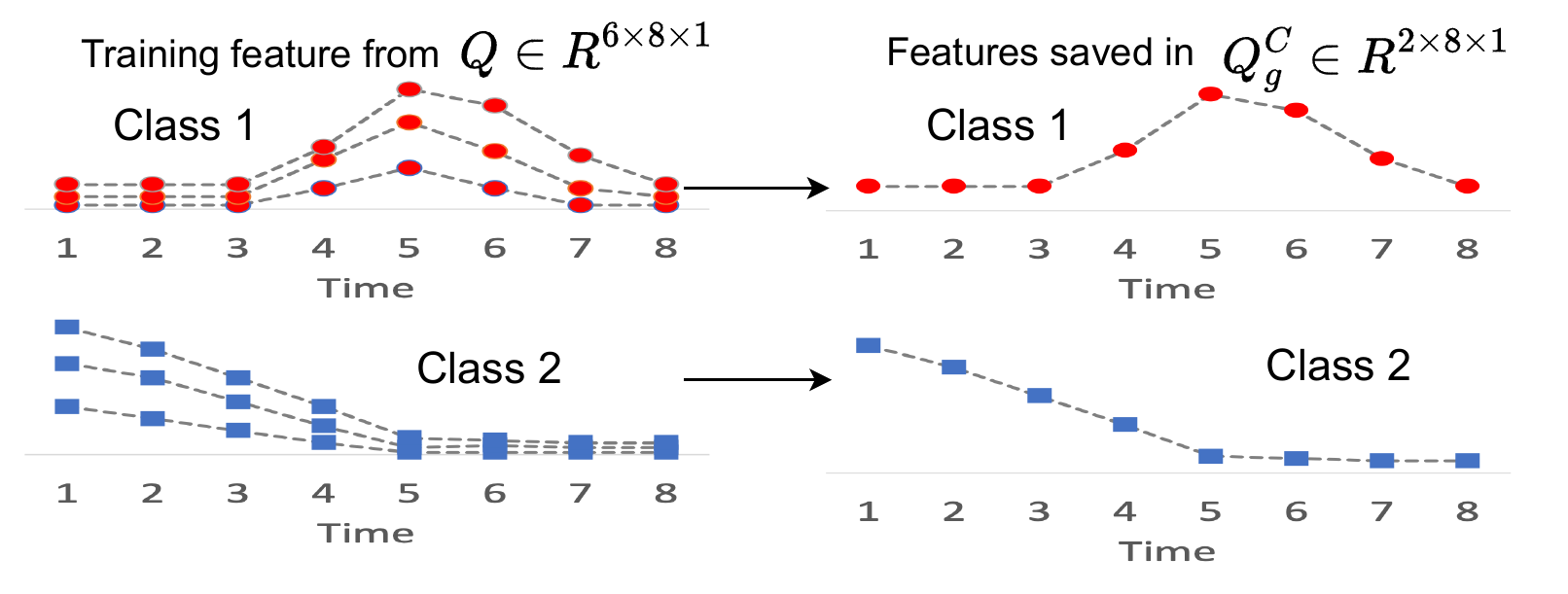}\\
\caption{Example from $\mathcal{Q}$ to $\mathcal{Q}^C$ (The points with the same 
color
represent the features for one specific class)}
\label{fig:QtoQC}
\end{center}
\end{figure}

Using the similar  mechanism to learn  
attention 
models~\cite{ref:first_attn_iclr2015, ref:google_all_atte_nips2017, ref:cross_attn_ijcai2020}, 
we learn the hidden features in query space ($\mathcal{Q}$) and key space ($\mathcal{K}$). Then, the class-specific features ($\mathcal{Q}^C$ and $\mathcal{K}^C$) are generated as follows.
From $\mathcal{Q}$ (or $\mathcal{K}$), the algorithm  finds instances from a specific class label $c$ and averages their features to be $\mathcal{Q}^C$ (or $\mathcal{K}^C$) (Line~\ref{csa_cal:class_key}-\ref{csa_cal:class_query}). The first dimension of $\mathcal{Q}^C$ is 
not the number of instances ($N$) as that in $\mathcal{Q}$. It becomes 
the number of classes ($C$).
This process can be demonstrated using an example in Fig.~\ref{fig:QtoQC}. 
It 
explains how  
the features specific to class 1 and class 2 are calculated. 

\subsubsection{Learning Process for $\mathbb{A}^C$ }
The calculation of the class-specific attention $\mathbb{A}^C$ utilizes the class-specific query features $\mathcal{Q}^C$ and the key features $\mathcal{K}^C$. 

To calculate the 
similarity between $\mathcal{Q}^C$ and $\mathcal{K}^C$, 
a feature matrix is calculated as 
$\mathcal{S} = \mathcal{K}^C\cdot (\mathcal{Q}^C)^T$. 
Different from existing attention mechanisms, we introduce a class differentiation (CD) component to post-process the feature matrix $\mathcal{S}$, as shown as the right shaded area in Fig.~\ref{fig:csa_details}. 
Existing attention mechanisms calculate the attention values directly from $\mathcal{S}$.  However, 
calculating the similarity between the key and query features is not 
sufficient
for CSA. If a feature is learned from instances in multiple (or all) classes,  finding this feature may not be very helpful when making predictions. To better separate instances from one class, CSA needs to identify specific features from instances within 
this class, which are not commonly found from instances belonging to other classes.
By adding the absolute differences between features for class $c$ and all the other classes ($\neg c$) (Lines~\ref{csa_cal:s_diff1}-\ref{csa_cal:s_diff2}), 
we seek features that can 
differentiate instances from class $c$ and instances from other classes ($\neg c$) 
while still making use of the original features.
%
The significant class-specific features are kept in $\mathcal{S}^C$. 
$\mathcal{S}^C$ is further used to calculate the class-specific attention $\mathbb{A}^C$ (Line~\ref{csa_cal:a_final}).

\subsubsection{Updating of $\mathcal{L}$ to $\mathcal{O}_{CSA}$}
Class-specific attention $\mathbb{A}^C$ is used to adjust the value features $\mathcal{V}$
from $\mathcal{L}$ 
to get $\mathcal{O}_{CSA}$ (Lines~\ref{csa_cal:value}-\ref{csa_cal:return}). 

The value features $\mathcal{V}$ and the features in $\mathcal{L}$ are transformed to $\mathcal{V}^C$ and $\mathcal{L}^C$ by repeat $C$ copies. 
The output features in $\mathcal{O}_{CSA}$ are calculated using $\mathcal{O}_{CSA}=\mathcal{L}^C + \sigma\times (\mathbb{A}^C \cdot \mathcal{V}^C)$ where $\sigma$ is a learnable scalar value. 
\subsection{
Utilization of CSA for the Output of NN
}
In the {\em CSA-embedded FCN} model (i.e., FCN model that embeds the CSA module, as shown in Figure~\ref{fig:csa_bigpicture}), the features in $\mathcal{O}_{CSA}$ are adjusted using class-specific attention and are fed to output layers (global pooling layers and fully connected layers).
More specifically, the output layers of the {\em  CSA-embedded FCN} consist of one global pooling layer and one fully connected layer. The design of the global pooling layer follows the common design practice in existing models~\cite{ref:lstm_fcn_acc2018,ref:mlstm_fcn_nn2019,ref:fcn_ijcnn2017}:
$\mathcal{O}_{CSA}$ ($\in R^{N\!\times\!C\!\times\!T\!\times\!F}$) is converted to a condensed feature tensor $\mathcal{G}\!\in\! R^{N\!\times\!C\!\times\!F}$ by averaging the features in the time dimension. 
%


%
%

\begin{table}[b!]
\centering
\small
\begin{tabular}{l|| c | c | c | c }
\toprule
Dataset & $N$ & $C$ & $V$ & T \\
\midrule
ArtWordRec      & 575   & 25  & 9   & 144   \\
BasicMotions    & 80    & 4   & 6   & 100   \\
CharTraj        & 2858  & 20  & 3   & 182   \\
Cricket         & 180   & 12  & 6   & 1197  \\
DuckDuckGeese   & 100   & 5   & 270 & 1345  \\
EigenWorms      & 259   & 5   & 6   & 17984 \\
Epilepsy        & 275   & 4   & 3   & 206   \\
EthanolConc     & 1751  & 4   & 3   & 1751  \\
FaceDetection   & 9414  & 2   & 62  & 144   \\
FingerMovements & 416   & 2   & 28  & 50    \\
HandMovement    & 234   & 4   & 10  & 400   \\
Handwriting     & 1000  & 26  & 3   & 152   \\
Heartbeat       & 409   & 2   & 61  & 405   \\
InsectWingbeat  & 50000 & 10  & 30  & 200   \\
JapaneseVowels  & 640   & 10  & 26  & 12    \\
LSST            & 4925  & 14  & 6   & 36    \\
Libras          & 360   & 15  & 2   & 45    \\
MotorImagery    & 378   & 2   & 64  & 3000  \\
NATOPS          & 360   & 6   & 24  & 51    \\
PEMS-SF         & 440   & 7   & 963 & 144   \\
PenDigits       & 10992 & 10  & 2   & 8     \\
Phoneme         & 6668  & 39  & 11  & 217   \\
RacketSports    & 303   & 4   & 6   & 30    \\
SelfRegSCP1     & 561   & 2   & 6   & 896   \\
SelfRegSCP2     & 380   & 2   & 7   & 1152  \\
SpokenArab      & 8798  & 10  & 13  & 93    \\
StandWalkJump   & 27    & 3   & 4   & 2500  \\
UWaveGesture    & 440   & 8   & 3   & 315   \\\hline
MedicalImages                 &  1141  &  10   &  1  & 99     \\
MelPedestrian                 &  3633  &  10   &  1  & 24     \\
MidPhalanxoutGrp              &  554   &  3    &  1  & 80     \\
MidPhalanxoutCor              &  891   &  2    &  1  & 80     \\
MidPhalanxtw                  &  553   &  6    &  1  & 80     \\
OsuLeaf                       &  442   &  6    &  1  & 427    \\
PhalangesCor                  &  2658  &  2    &  1  & 80     \\
Powercons                     &  360   &  2    &  1  & 144    \\
ProximalPhaGrp                &  605   &  3    &  1  & 80     \\
ProximalPhaCor                &  891   &  2    &  1  & 80     \\
ProximalPhaTw                 &  605   &  6    &  1  & 80     \\
RefrigerationDev              &  750   &  3    &  1  & 720    \\

\bottomrule
\end{tabular}
\vspace{0.05in}
\caption{Statistics of 28 MTS and 12 UTS Datasets}
\label{tb:data}
\end{table}

{\em Where CSA differs from the conventional design} is in the fully connected layer: the features in $\mathcal{G}$, if directly fed to a traditional fully connected layer, cannot make good use of the features from specific classes because fully-connected layers
directly combine all features without considering their differences.
%
Instead, we design a layer 
by introducing class-specific weights and biases to better utilize class-specific features. 
In particular, for each class $c$, a weight matrix 
$\Omega \in R^{F\!\times\!1}$
and a bias $\beta$ 
are used to 
calculate a value 
for this class 
($val=\mathcal{G}\cdot \Omega\! +\!\beta$). This value is 
used to get 
 the probability of an instance belonging to class $c$. 
%
Compared with the traditional fully-connected layer,
this design can make better use of the class-specific features from $\mathcal{O}_{CSA}$ because features from different classes
are not combined. Instead, the probability of an instance belonging to one class is calculated only using features specific to this class for most cases. Meanwhile, the weight parameters (in $\Omega$ and $\beta$) are not more than those in a traditional fully-connected layer -- therefore, the training time for this fully-connected layer does not increase.

{\bf Prediction for a testing instance.} 
The class-specific features $\mathcal{O}_{CSA}$ for a testing instance are calculated by directly utilizing the global  class-specific attention $\mathbb{A}^C$ and the value features of this instance.
Note that the testing instance does not need any class label to leverage class-specific attention.
Next, these class-specific features $\mathcal{O}_{CSA}$ are passed to the specially designed fully connected layer described above to make class predictions.

\section{Experiments}
\label{sec:exp}
In this section, we evaluate the effectiveness of the proposed CSA module. 
In particular, we aim to answer the following questions: \\
(Q1) Does the CSA module improve the performance of a baseline {\em neural network} model (which may or may not implement a general attention mechanism) to classify time series data?\\
(Q2) Does the CD component contribute to improving the performance of the CSA module? and\\ 
(Q3) Will the CSA module introduce dramatic overheads (regarding training time) in the model learning process?

All  methods 
are implemented using $Python~3.7$ and tested on a server with Intel Xeon Gold 5218 2.3G CPUs, 192GB RAM, and one Nvidia Tesla V100 GPU with 32GB memory. PyTorch 1.10 
is used to build all the  models. To acquire stable results, every number (accuracy or running time) we report is an {\bf average of five} runs.

\subsection{Experimental Settings}
{\bf Datasets.} 
We report results on 40 datasets, where 28 MTS datasets are from the UEA repository~\cite{ref:uea_data} and 12 UTS datasets are from the UCR repository~\cite{ref:ucr_data}. 
Two datasets from the UEA are not utilized because they have too few instances and the UTS datasets are randomly chosen.

%
%

\noindent{\bf Methods for Comparison. }
The CSA module is designed to be compatible with, and improve the time series classification performance of, existing neural network (NN) models. 
Many neural network models exist to classify time series, it is not realistic to embed CSA to all of them for testing. 
%
In our experiments, we embed the CSA module in five {\em representative} state-of-the-art models, namely
(1) Fully Convolutional Networks ({\em FCN})~\cite{ref:fcn_ijcnn2017}, 
(2) Multivariate Long Short-Term Memory ({\em MLSTM})~\cite{ref:lstm_neco1997}, 
(3) {\em MLSTM-FCN}~\cite{ref:mlstm_fcn_nn2019},  
(4) CNN with attention ({\em CNN-ATN})~\cite{ref:cross_attn_ijcai2020}, 
and (5) {\em TapNet}~\cite{ref:tapnet_aaai2020}. The first three models are widely used NN models without attention mechanism, while {\em CNN-ATN} leverages attention in MTS classification, {\em TapNet} incorporates both attention and  class label information in training. 
%
%
For these five models, we get their implementation from the published source code. If the code version is not comparable, we convert the code 
to PyTorch. 
%
Recall that the CSA module implements a CD component. To evaluate the effectiveness of the CD component of the CSA module, we implement a CSA module without the CD component for testing.
%


\noindent {\bf Hyper-Parameter Setting. }
We use the hyper-parameter settings reported in
\cite{ref:mlstm_fcn_nn2019}, and~\cite{ref:cross_attn_ijcai2020}:
the convolutional layers of the {\em FCN} model contain three 1D kernels with sizes {8, 5, 3} -- the corresponding numbers of kernels are 128, 256, and 128. The pooling filter in the global pooling layer has the same as the length as the time series output from the previous layer. The 2D kernels in the {\em CNN-ATN} model have sizes {($8\times 1$), ($5\times 1$), and ($3\times 1$)}. 
The number of the hidden states
of {\em MLSTM} model is 128.
The $F$ parameter for  $\mathcal{L}$ (Fig.~\ref{fig:csa_details}) for all the models is set to 128. For TapNet, the number of variable subsets is set to  3. 
We do not test the effect of these hyper-parameters because they are intrinsic in the base models, but not the CSA module. 
Tuning the parameters to get good  performance is not the focus of this paper. We use settings that are the same or similar to their original papers.

\noindent{\bf Evaluation Measure.} We 
report
a commonly used classification performance measure, accuracy ($Acc \in [0, 1]$), because most datasets are balanced. 
We also compute and report 
the accuracy improvement, $AI$, 
 to compare two models, $A$ and $B$, with accuracy values $Acc_A$ and $Acc_B$:
 \vspace{0.05in}
 \begin{equation}
     AI(A,B)
     = \frac{Acc_A-Acc_B}{Acc_B}
     \label{eq:ai}
 \end{equation}
 \vspace{0.05in}
Since we are generally interested in measuring the improvement provided by the CSA module over the conventional baselines, we use $AI_{\text{\em Model}}$ to denote $AI(\text{{\em Model-CSA}},\text{{\em Model}})$.
Each experiment were run {\bf five} times on each dataset and the average accuracy and improvements for each dataset are reported. 

We also conduct statistical analysis, the null hypothesis testing~\cite{ref:null_hypothesis}, 
to gain a more in-depth understanding of the effectiveness of the CSA module.
Given one dataset, the null hypothesis is that the {\em Model-CSA} performs similarly with (not significantly better than) the base {\em Model}. The $p$-value is calculated using 
the Chi-Square Test~\cite{ref:pearson_chisquare}. If $p$-value is bigger than a threshold (a commonly used threshold is 0.05), the null hypothesis is rejected and we consider the model with higher accuracy is significantly better than the other model.


\subsection{Effectiveness Analysis} 
This section presents the effectiveness of the CSA module.

\subsubsection{MTS datasets}
We first study the effectiveness of plugging CSA to a neural network model on classifying MTS datasets.
We report the detailed accuracy/{\em AI} values in 
Table~\ref{tb:mts_details}. 
For each model, the first two columns (`{\em w/o CSA}' and `{\em w CSA}' show the accuracy of the models and the third column (`{\em AI}') shows the accuracy improvement of {\em Model-CSA} over the base {\em Model}. Table~\ref{tb:uts_details} also uses this convention.

In this table, the results on several datasets for {\em TapNet} and {\em CNN-ATN}  model are missing (denoted with '-') because these models cannot finish utilizing the GPU memory.
{\em CNN-ATN} uses 2D convolutional operations, as well as attentions over variable and temporal dimensions, which requires a large memory space. 
{\em TapNet} generates an $\mathcal{L}$ tensor from three subsets of the input MTS data, thus utilizes much more space.

\begin{table*}[ht!]
\small
\begin{subtable}[t]{\textwidth}
\centering
\begin{tabular}{ l ||| c |  c || c ||| c | c || c ||| c | c || c}
\toprule
 & \multicolumn{3}{c|||}{{\text{{\em FCN}}}}
 & \multicolumn{3}{c|||}{{\text{{\em MLSTM}}}}
 & \multicolumn{3}{c}{{\text{{\em MLSTM-FCN}}}}
 \\\cline{2-10}
 Datasets  & {\text{{\em w/o CSA}}} & {\text{{\em w CSA}}} & {\text{{\em AI}}} 
 & {\text{{\em w/o CSA}}} & {\text{{\em w CSA}}} & {\text{{\em AI}}} 
 & {\text{{\em w/o CSA}}} & {\text{{\em w CSA}}} & {\text{{\em AI}}} \\
\midrule
ArtWordRec       & 0.980 & 0.982 & \bf{0.204 } & 0.816 & 0.856 &  \bf{4.902 } & 0.972 & 0.982 & \bf{1.029 } \\
BasicMotions     & 0.968 & 0.966 &    -0.207   & 0.844 & 0.852 &  \bf{0.948 } & 0.966 & 0.968 & \bf{0.207 } \\
CharTraj         & 0.990 & 0.990 &     0.000   & 0.974 & 0.980 &  \bf{0.616 } & 0.992 & 0.996 & \bf{0.403 } \\
Cricket          & 0.910 & 0.910 &     0.000   & 0.704 & 0.708 &  \bf{0.568 } & 0.878 & 0.896 & \bf{2.050 } \\
DuckDuckGeese    & 0.740 & 0.766 & \bf{3.514 } & 0.672 & 0.694 &  \bf{3.274 } & 0.722 & 0.734 & \bf{1.662 } \\
EigenWorms       & 0.544 & 0.552 & \bf{1.471 } & 0.464 & 0.464 &      0.000   & 0.526 & 0.574 & \bf{9.125 } \\
Epilepsy         & 0.794 & 0.842 & \bf{6.045 } & 0.666 & 0.686 &  \bf{3.003 } & 0.864 & 0.874 & \bf{1.157 } \\
EthanolConc      & 0.624 & 0.658 & \bf{5.449 } & 0.342 & 0.346 &  \bf{1.170 } & 0.622 & 0.654 & \bf{5.145 } \\
FaceDetection    & 0.562 & 0.562 &     0.000   & 0.578 & 0.582 &  \bf{0.692 } & 0.556 & 0.562 & \bf{1.079 } \\
FingerMovements  & 0.648 & 0.656 & \bf{1.235 } & 0.614 & 0.614 &      0.000   & 0.642 & 0.642 &     0.000   \\
HandMovement     & 0.468 & 0.488 & \bf{4.274 } & 0.370 & 0.414 &  \bf{11.892} & 0.466 & 0.492 & \bf{5.579 } \\
Handwriting      & 0.284 & 0.288 & \bf{1.408 } & 0.208 & 0.224 &  \bf{7.692 } & 0.282 & 0.296 & \bf{4.965 } \\
Heartbeat        & 0.812 & 0.818 & \bf{0.739 } & 0.818 & 0.820 &  \bf{0.244 } & 0.810 & 0.816 & \bf{0.741 } \\
InsectWingbeat   & 0.108 & 0.138 & \bf{27.778} & 0.124 & 0.124 &      0.000   & 0.116 & 0.136 & \bf{17.241} \\
JapaneseVowels   & 0.882 & 0.890 & \bf{0.907 } & 0.938 & 0.952 &  \bf{1.493 } & 0.932 & 0.934 & \bf{0.215 } \\
LSST             & 0.444 & 0.456 & \bf{2.703 } & 0.528 & 0.530 &  \bf{0.379 } & 0.530 & 0.558 & \bf{5.283 } \\
Libras           & 0.902 & 0.906 & \bf{0.443 } & 0.552 & 0.708 &  \bf{28.261} & 0.906 & 0.902 &     -0.442  \\
MotorImagery     & 0.640 & 0.648 & \bf{1.250 } & 0.608 & 0.624 &  \bf{2.632 } & 0.638 & 0.638 &     0.000   \\
NATOPS           & 0.886 & 0.898 & \bf{1.354 } & 0.792 & 0.820 &  \bf{3.535 } & 0.882 & 0.898 & \bf{1.814 } \\
PEMS-SF          & 0.930 & 0.946 & \bf{1.720 } & 0.576 & 0.704 &  \bf{22.222} & 0.928 & 0.944 & \bf{1.724 } \\
PenDigits        & 0.984 & 0.982 &     -0.203  & 0.976 & 0.976 &      0.000   & 0.980 & 0.980 &     0.000   \\
Phoneme          & 0.088 & 0.092 & \bf{4.545 } & 0.100 & 0.106 &  \bf{6.000 } & 0.096 & 0.114 & \bf{18.750} \\
RacketSports     & 0.748 & 0.774 & \bf{3.476 } & 0.778 & 0.790 &  \bf{1.542 } & 0.770 & 0.804 & \bf{4.416 } \\
SelfRegSCP1      & 0.872 & 0.874 & \bf{0.229 } & 0.898 & 0.904 &  \bf{0.668 } & 0.884 & 0.888 & \bf{0.452 } \\
SelfRegSCP2      & 0.568 & 0.586 & \bf{3.169 } & 0.572 & 0.596 &  \bf{4.196 } & 0.592 & 0.592 &     0.000   \\
SpokenArab       & 0.974 & 0.982 & \bf{0.821 } & 0.960 & 0.960 &      0.000   & 0.982 & 0.986 & \bf{0.407 } \\
StandWalkJump    & 0.400 & 0.400 &     0.000   & 0.692 & 0.652 &      -5.780  & 0.428 & 0.414 &     -3.271  \\
UWaveGesture     & 0.624 & 0.618 &     -0.962  & 0.814 & 0.820 &  \bf{0.737 } & 0.768 & 0.766 &     -0.260  \\\hline
Averaged  & & & \bf{2.549} & & & \bf{3.603} & & & \bf{2.838}\\
\bottomrule  
 \end{tabular}
\end{subtable}\vspace{0.05in}\\
    \hfill
\begin{subtable}[t]{\textwidth}
\centering
\begin{tabular}{ l ||| c |  c || c  ||| c | c || c }
\toprule
 & \multicolumn{3}{c|||}{{\text{{\em TapNet}}}}
 & \multicolumn{3}{c}{{\text{{\em CNN-ATN}}}}
 \\\cline{2-7}
 Datasets  & {\text{{\em w/o CSA }}} & {\text{{\em w CSA}}}  & {\text{{\em AI}}} & {\text{{\em w/o CSA}}} & {\text{{\em w CSA}}} & {\text{{\em AI}}} \\
\midrule
ArtWordRec       & 0.980 & 0.988 & \bf{0.816}  &  0.984  &  0.988  &  \bf{0.407} \\
BasicMotions     & 1.000 & 0.994 &    -0.600   &  0.982  &  1.000  &  \bf{1.833} \\
CharTraj         & 0.990 & 0.990 &     0.000   &  0.990  &  0.990  &  0.000      \\
Cricket          & 0.950 & 0.932 &    -1.895   &  0.808  &  0.826  &  \bf{2.228} \\
DuckDuckGeese    & 0.724 & 0.676 &    -6.630   &  -      &  -      &  -          \\
EigenWorms       & 0.600 & 0.580 &    -3.333   &  -      &  -      &  -          \\
Epilepsy         & 0.870 & 0.822 &    -5.517   &  0.950  &  0.954  &  \bf{0.421} \\
EthanolConc      & 0.322 & 0.364 & \bf{13.043} &  0.520  &  0.510  &  -1.923     \\
FaceDetection    & 0.562 & 0.570 & \bf{1.423}  &  -      &  -      &  -          \\
FingerMovements  & 0.614 & 0.636 & \bf{3.583}  &  0.602  &  0.620  &  \bf{2.990} \\
HandMovement     & 0.464 & 0.520 & \bf{12.069} &  0.422  &  0.442  &  \bf{4.739} \\
Handwriting      & 0.198 & 0.278 & \bf{40.404} &  0.296  &  0.298  &  \bf{0.676} \\
Heartbeat        & 0.748 & 0.732 &    -2.139   &  0.820  &  0.820  &  0.000      \\
InsectWingbeat   & -     & -     &     -       &  -      &  -      &  -          \\
JapaneseVowels   & 0.988 & 0.992 & \bf{0.405 } &  0.990  &  0.990  &  0.000      \\
LSST             & 0.450 & 0.642 & \bf{42.667} &  0.664  &  0.674  &  \bf{1.506} \\
Libras           & 0.790 & 0.820 & \bf{3.797}  &  0.790  &  0.790  &  0.000      \\
MotorImagery     & 0.636 & 0.614 &    -3.459   &  -      &  -      &  -          \\
NATOPS           & 0.890 & 0.890 &     0.000   &  0.946  &  0.950  &  \bf{0.423} \\
PEMS-SF          & 0.912 & 0.912 &     0.000   &  -      &  -      &  -          \\
PenDigits        & 0.940 & 0.940 &     0.000   &  0.970  &  0.970  &  0.000      \\
Phoneme          & 0.164 & 0.216 & \bf{31.707} &  -      &  -      &  -          \\
RacketSports     & 0.704 & 0.816 & \bf{15.909} &  0.888  &  0.888  &  0.000      \\
SelfRegSCP1      & 0.736 & 0.636 &    -13.587  &  0.830  &  0.878  &  \bf{5.783} \\
SelfRegSCP2      & 0.578 & 0.590 & \bf{2.076}  &  0.546  &  0.558  &  \bf{2.198} \\
SpokenArab       & 0.982 & 0.982 &     0.000   &  0.990  &  0.990  &  0.000      \\
StandWalkJump    & 0.508 & 0.588 & \bf{15.748} &  -      &  -      &  -          \\
UWaveGesture     & 0.882 & 0.874 &    -0.907   &  0.838  &  0.834  &  -0.477     \\\hline
Averaged & & & \bf{5.392} & & & \bf{1.040} \\
\bottomrule
\end{tabular}
\end{subtable}
\caption{
Comparison of models with and w/o CSA on MTS 
}
\label{tb:mts_details}
\end{table*}
\clearpage
\begin{table}[ht!]
\centering
\small
\begin{tabular}{l||c|c||c}
\toprule
& \multicolumn{3}{c}{{\em Model-CSA is Better}}
 \\\cline{2-4}
Base Model & Significantly & Not Significantly & Total\\ 
\midrule
{\em FCN} & 3/28 & 18/28 & 21/28 \\
{\em MLSTM} & 2/28 & 20/28 & 22/28 \\
{\em MLSTM-FCN} & 4/28 & 17/28 & 21/28 \\
{\em TapNet} & 5/27 & 8/27 & 13/27 \\
{\em CNN-ATN} & 1/20 & 10/20 & 11/20 \\\hline
{\em Percentages} & 15/131  & 73/131 & 88/131 \\
{\em of test cases} & =\bf{11.45\%} & =\bf{55.73\%} & =\bf{67.18\%} \\
\bottomrule
\end{tabular}
\vspace{0.05in}
\caption{Statistical significance analysis for all the models on MTS datasets ($p$-value threshold: 0.05; 
Each test case is for one dataset being applied with a base model and its CSA counterpart. }
\label{tb:mts_significant}
\end{table}

\begin{table*}[hb!]
\begin{subtable}[t]{\textwidth}
\small
\centering
\begin{tabular}{ l ||| c |  c || c ||| c | c || c ||| c | c || c}
\toprule
 & \multicolumn{3}{c|||}{{\text{{\em FCN }}}}
 & \multicolumn{3}{c|||}{{\text{{\em LSTM}}}}
 & \multicolumn{3}{c}{{\text{{\em LSTM-FCN}}}}
 \\\cline{2-10}
 Datasets  & {\text{{\em w/o CSA}}} & {\text{{\em w CSA}}} & {\text{{\em AI}}} & {\text{{\em w/o CSA}}} & {\text{{\em w CSA}}} & {\text{{\em AI}}} & {\text{{\em w/o CSA}}} & {\text{{\em w CSA}}} & {\text{{\em AI}}} \\
\midrule
MedicalImages 
                                &   0.792   &  0.794   & \bf{0.253}    & 0.522    & 0.580    & \bf{11.111}   & 0.786  & 0.790  & \bf{0.509}   \\
MelPedestrian 
                                &   0.660   &  0.666   & \bf{0.909}    & 0.830    & 0.900    & \bf{8.434}    & 0.796  & 0.798  & \bf{0.251}   \\
MidPhalanxoutGrp 
                                &   0.604   &  0.636   & \bf{5.298}    & 0.428    & 0.580    & \bf{35.514}   & 0.618  & 0.616  & -0.323       \\
MidPhalanxoutCor 
                                &   0.814   &  0.816   & \bf{0.246}    & 0.570    & 0.570    & 0.000         & 0.810  & 0.812  & \bf{0.247}   \\
MidPhalanxtw 
                                &   0.516   &  0.516   & 0.000         & 0.458    & 0.554    & \bf{20.961}   & 0.500  & 0.516  & \bf{3.200}   \\
OsuLeaf 
                                &   0.966   &  0.974   & \bf{0.828}    & 0.210    & 0.210    & 0.000         & 0.964  & 0.976  & \bf{1.245}   \\
PhalangesCor 
                                &   0.804   &  0.806   & \bf{0.249}    & 0.600    & 0.600    & 0.000         & 0.798  & 0.806  & \bf{1.003}   \\
Powercons 
                                &   0.926   &  0.930   & \bf{0.432}    & 0.940    & 0.952    & \bf{1.277}    & 0.926  & 0.932  & \bf{0.648}   \\
ProximalPhaGrp  
                                &   0.838   &  0.846   & \bf{0.955}    & 0.590    & 0.738    & \bf{25.085}   & 0.840  & 0.842  & \bf{0.238}   \\
ProximalPhaCor  
                                &   0.846   &  0.864   & \bf{2.128}    & 0.690    & 0.720    & \bf{4.348}    & 0.834  & 0.860  & \bf{3.118}   \\
ProximalPhaTw 
                                &   0.744   &  0.728   & -2.151        & 0.734    & 0.824    & \bf{12.262}   & 0.718  & 0.744  & \bf{3.261}   \\
RefrigerationDev  
                                &   0.520   &  0.516   & -0.769        & 0.408    & 0.464    & \bf{13.725}   & 0.528  & 0.536  & \bf{1.515}   \\\hline
Averaged     & &                & \bf{0.644} & &  & \bf{11.251} & &  & \bf{1.175} \\
\bottomrule
 \end{tabular}
\end{subtable}\vspace{0.05in}\\
    \hfill
\begin{subtable}[t]{\textwidth}
\small
\centering
\begin{tabular}{ l ||| c |  c || c  ||| c | c || c }
\toprule
 & \multicolumn{3}{c |||}{{\text{{\em TapNet}}}}
 & \multicolumn{3}{c}{{\text{{\em CNN-ATN}}}}
 \\\cline{2-7}
 Datasets  & {\text{{\em w/o CSA}}} & {\text{{\em w CSA}}} & {\text{{\em AI}}} & {\text{{\em w/o CSA}}} & {\text{{\em w CSA}}} & {\text{{\em AI}}} \\
\midrule
MedicalImages 
                                &  0.750  & 0.752  & \bf{0.267}  & 0.766   & 0.774   & \bf{1.044}\\
MelPedestrian 
                                &  0.792  & 0.822  & \bf{3.788}  & 0.850   & 0.862   & \bf{1.412}\\
MidPhalanxoutGrp 
                                &  0.658  & 0.662  & \bf{0.608}  & 0.668   & 0.682   & \bf{2.096}\\
MidPhalanxoutCor 
                                &  0.844  & 0.848  & \bf{0.474}  & 0.826   & 0.828   & \bf{0.242}\\
MidPhalanxtw 
                                &  0.618  & 0.610  &  -1.294     & 0.540   & 0.542   & \bf{0.370}\\
OsuLeaf 
                                &  0.972  & 0.982  & \bf{1.029}  & 0.898   & 0.916   & \bf{2.004}\\
PhalangesCor 
                                &  0.772  & 0.790  & \bf{2.332}  & 0.806   & 0.822   & \bf{1.985}\\
Powercons 
                                &  0.944  & 0.960  & \bf{1.695}  & 0.918   & 0.928   & \bf{1.089}\\
ProximalPhaGrp  
                                &  0.856  & 0.854  &  -0.234     & 0.850   & 0.852   & \bf{0.235}\\
ProximalPhaCor  
                                &  0.894  & 0.900  & \bf{0.671}  & 0.918   & 0.908   & -1.089\\
ProximalPhaTw 
                                &  0.794  & 0.800  & \bf{0.756}  & 0.818   & 0.828   & \bf{1.222}\\
RefrigerationDev  
                                &  0.584  & 0.584  &  0.000      & 0.582   & 0.568   & -2.405\\\hline
Averaged & & & \bf{0.776} & & &  \bf{0.631} \\
\bottomrule
\end{tabular}
\end{subtable}
\caption{
Comparison of models with and w/o CSA on UTS
}
\label{tb:uts_details}
\end{table*}

The table clearly show that the {\em AI} values is bigger than zero for most datasets for each base model. This means that the CSA module helps improve all these different base models.
The average {\em AI} for the different models ranges from 1.04\% to 5.39\%. 
%
There are a few cases where the base models show slightly better performance than their corresponding CSA model; we conjecture that this is due to the increase in the number of model parameters when using the CSA extension. 
The accuracy improvement over the {\em TapNet} method is the highest. This is because {\em TapNet} generates 
the $\mathcal{L}$ tensor from three subsets of the input MTS data, which is three times larger than the $\mathcal{L} \in R^{N\times T\times F}$ from other models.
%
The large $\mathcal{L}$ provides more information for our CSA module to utilize.

The statistical analysis is conducted on all 28 MTS datasets. 
Table~\ref{tb:mts_significant} presents the results.
The first five rows present the percentage of datasets and the last row shows the percentage of all test cases, where each test case is for one dataset being applied with a base model and its CSA counterpart. 
There are $28$ MTS datasets with 5 different base NN models. Ideally, the total number of test cases is $28\times 5 = 140$. However, {\em TapNet} can only finish 27 datasets and {\em CNN-ATN} can only finish 20 datasets. 
%
The total number of available test cases is 131 ($28+28+28+27+20$).
Among all the 131 cases, in  $11.45\%$ (15/131) cases, the {\em Model-CSA} significantly outperform the base {\em Model}. 
The {\em Model-CSA} is better but not statistically significant better than the {\em Model} in $55.73\%$ cases. Overall, {\em Model-CSA} is better than {\em Model} in 
$67.18\%$ cases.
\subsubsection{UTS datasets} 
Table~\ref{tb:uts_details} presents the accuracy/{\em AI} values on UTS for all the models.
The results show that the proposed CSA module can also improve the classification performance of all the base models on UTS datasets. The  average  {\em AI}  for  the  different  models  ranges  from 0.6\% to 11\%
Compared with the results on the MTS datasets, the {\em AI} results are slightly smaller on UTS datasets. This is because MTS datasets normally have more features than the UTS datasets. 

\begin{table}[htp]
\centering
\small
\begin{tabular}{l|||c|c||c}
\toprule
& \multicolumn{3}{c}{{\em Model-CSA is Better}}
 \\\cline{2-4}
Base Model & Significantly & Not Significantly & Total\\ 
\midrule
{\em FCN} & 0/12 & 9/12 & 9/12 \\
{\em LSTM} & 7/12 & 2/12 & 9/12 \\
{\em LSTM-FCN} & 0/12 & 11/12 & 11/12 \\
{\em TapNet} & 1/12 & 8/12 & 9/12 \\
{\em CNN-ATN} & 0/12 & 10/12 & 10/12 \\\hline
{\em Percentages} & 8/60  & 40/60 & 48/60 \\
{\em of test cases} & \bf{=13.33\%} & \bf{=66.67\%} & \bf{=80.00\%} \\
\bottomrule
\end{tabular}
\vspace{0.05in}
\caption{Statistical significance analysis for all the models on UTS datasets ($p$-value threshold: 0.05; Each test case is for one dataset being applied with a base model and its CSA counterpart)}
\label{tb:uts_significant}
\end{table}

We want to particular mention that the improvement on the {\em TapNet} model is low on UTS. The reason is that there is only one variable in a UTS. Thus, we cannot create multiple variable subsets to generate $\mathcal{L}$. 
The best improvement is observed on the {\em LSTM} model (which corresponds to the {\em MLSTM} model for MTS). This is because {\em LSTM} is the worst performing base model on UTS, which can be observed from the accuracy values in the {\em `LSTM w/o CSA'} column .

We also conduct the null hypothesis test on the UTS datasets. We run the tests on all the 12 UTS datasets and present the results 
in 
Table~\ref{tb:uts_significant}.
Again, since we test on five models, there are 60 test cases ($5\times 12$). 
In about $13.33\%$ cases, a model with CSA significantly outperforms 
the base model and, in $66.67\%$ cases, there is improvement (although not statistically significant). 
The total percentage of benefited cases from the CSA module is $80\%$.
On UTS datasets, the majority improvements are from the {\em LSTM} model, it might be because that the {\em LSTM} model has relatively lower performance on the 12 UTS datasets. 

Similar with the conclusion on MTS datasets, {\em Model-CSA}
reports 
better 
performance than the base {\em Model} 
on UTS datasets.
In the rest of the experimental sections, we 
present the results from tests on the MTS datasets to avoid redundant contents.

\subsection{Ablation Study}
\begin{table}[htb]
\centering
\small
\begin{tabular}{ l ||| c |  c || c }
\toprule
 & \multicolumn{3}{c}{{\text{{\em FCN-CSA}}}}
 \\\cline{2-4}
Datasets  & {\text{{\em w/o CD}}} & {\text{{\em w CD}}} & {\text{{\em AI}}} \\
\midrule
ArtWordRec      & 0.980 & 0.982 & \bf{0.204} \\
BasicMotions    & 0.962 & 0.966 & \bf{0.416} \\
CharTraj        & 0.990 & 0.990 & 0.000      \\
Cricket         & 0.914 & 0.910 & -0.438     \\
DuckDuckGeese   & 0.742 & 0.766 & \bf{3.235} \\
EigenWorms      & 0.562 & 0.552 & -1.780     \\
Epilepsy        & 0.818 & 0.842 & \bf{2.997} \\
EthanolConc     & 0.658 & 0.658 & 0.000      \\
FaceDetection   & 0.560 & 0.562 & \bf{0.357} \\
FingerMovements & 0.653 & 0.656 & \bf{0.536} \\
HandMovement    & 0.475 & 0.488 & \bf{2.737} \\
Handwriting     & 0.280 & 0.288 & \bf{2.857} \\
Heartbeat       & 0.813 & 0.818 & \bf{0.677} \\
InsectWingbeat  & 0.134 & 0.138 & \bf{2.985} \\
JapaneseVowels  & 0.890 & 0.890 & 0.000      \\
LSST            & 0.478 & 0.456 & -4.50      \\
Libras          & 0.896 & 0.906 & \bf{1.116} \\
MotorImagery    & 0.642 & 0.648 & \bf{0.935} \\
NATOPS          & 0.898 & 0.898 & 0.000      \\
PEMS-SF         & 0.943 & 0.946 & \bf{0.371} \\
PenDigits       & 0.980 & 0.982 & \bf{0.204} \\
Phoneme         & 0.086 & 0.092 & \bf{6.977} \\
RacketSports    & 0.764 & 0.774 & \bf{1.309} \\
SelfRegSCP1     & 0.886 & 0.874 & -1.354     \\
SelfRegSCP2     & 0.573 & 0.586 & \bf{2.358} \\
SpokenArab      & 0.972 & 0.982 & \bf{1.029} \\
StandWalkJump   & 0.418 & 0.400 & -4.192     \\
UWaveGesture    & 0.612 & 0.618 & \bf{0.980} \\\hline
Averaged         &       &       & \bf{0.715} \\
Wins            &       &       & \bf{19/28} \\
\bottomrule
\end{tabular}
\vspace{0.05in}
\caption{Comparison for {\em FCN-CSA} models with and w/o CD component on MTS datasets}
\label{tb:mts_CD_details}
\end{table}
CSA relies on a class-differentiation (CD) component to detect the features that can differentiate a class $c$ and the remaining ones $\neg c$. 
In order to assess the impact of the CD component, we implement a version of CSA without CD, {\em Model-CSA-NoCD} as follows. 
We calculate the CSA outputs by applying the {\em SoftMax} function directly on $\mathcal{S}$ to calculate $\mathcal{A}^C$, without calculating the feature differences between class $c$ and classes $\neg c$ (i.e., without computing  $\mathcal{S}^C_c$ and $\mathcal{S}^C$ in Fig.~\ref{fig:csa_details}). 
Without loss of generality, we choose one base model ({\em FCN model}) and run {\em FCN-CSA} and {\em FCN-CSA-NoCD}  over all the MTS datasets. 
As we expect, for most data sets, the CD component helps improve the performance.
As  shown in Table~\ref{tb:mts_CD_details},
on average, {\em FCN-CSA} increase the accuracy of {\em FCN-CSA-NoCD} with 0.72\%. Despite that this number is not big, we note that among all the MTS datasets, {\em FCN-CSA} outperforms {\em FCN-CSA-NoCD} on 19 datasets and they have same performance on 4 datasets. These results show that the CD component helps improve the time series classification performance in general.

\begin{table}[htb]
\centering
\small
\begin{tabular}{l|| c | c }
\toprule
& {\em K Nearest Neighbor } & \\
Dataset & {\em classifier with DTW$_d$} & {\em MLSTM-FCN-CSA} \\
\midrule
ArtWordRec      	 &  \bf{0.987} 	& 0.982	 		\\
BasicMotions    	 &  \bf{0.975} 	& 0.968	 		\\
CharTraj        	 &  0.99 		& \bf{0.996}	\\
Cricket         	 &  \bf{1} 		& 0.896	 		\\
DuckDuckGeese   	 &  0.6 		& \bf{0.734}	\\
EigenWorms      	 &  \bf{0.618} 	& 0.574	 		\\
Epilepsy        	 &  \bf{0.964} 	& 0.874	 		\\
EthanolConc     	 &  0.323 		& \bf{0.654}	\\
FaceDetection   	 &  0.529 		& \bf{0.562}	\\
FingerMovements 	 &  0.53  		& \bf{0.642}	\\
HandMovement    	 &  0.231 		& \bf{0.492}	\\
Handwriting     	 &  \bf{0.607} 	& 0.296	 		\\
Heartbeat       	 &  0.717 		& \bf{0.816}	\\
InsectWingbeat  	 &  0.12		& \bf{0.136}	\\
JapaneseVowels  	 &  \bf{0.949} 	& 0.934	 		\\
LSST            	 &  \bf{0.872} 	& 0.558	 		\\
Libras          	 &  0.551 		& \bf{0.902}	\\
MotorImagery    	 &  0.5   		& \bf{0.638}	\\
NATOPS          	 &  0.883 		& \bf{0.898}	\\
PEMS-SF         	 &  0.711 		& \bf{0.944}	\\
PenDigits       	 &  0.977 		& \bf{0.98}		\\
Phoneme         	 &  \bf{0.151} 	& 0.114	 		\\
RacketSports    	 &  0.803 		& \bf{0.804}	\\
SelfRegSCP1     	 &  0.775 		& \bf{0.888}	\\
SelfRegSCP2     	 &  0.539 		& \bf{0.592}	\\
SpokenArab      	 &  0.963 		& \bf{0.986}	\\
StandWalkJump   	 &  0.2 		& \bf{0.414}	\\
UWaveGesture    	 &  \bf{0.903} 	& 0.766			\\\hline
Averaged              &   0.677      & \bf{0.715}    \\
Wins                 &              & \bf{18/28}    \\
\bottomrule
\end{tabular}
\vspace{0.05in}
\caption{Accuracy comparison between {\em K nearest neighbor} classifier (with {\em DTW$_d$} for distance calculation) and {\em MLTSM-FCN-CSA} on MTS datasets}
\label{tb:dtw_fcnmlstm}
\end{table}

\subsection{Case Study: Comparison with non-NN methods}

The propose CSA module can help improve the performance of neural network-based models, but cannot 
be leveraged in the traditional time-series classification methods, such as the K-Nearest-Neighbor classifier~\cite{ref:knn}, Dynamic Time Warping ({\em DTW})~\cite{ref:dtw}.
The K-Nearest-Neighbor approach has shown excellent performance in classifying time series~\cite{DBLP:journals/pvldb/DingTSWK08,ref:uea_data}. 
In particular, 1-Nearest-Neighbor classifier with dimension-dependent {\em DTW} ({\em DTW$_d$})~\cite{ref:uea_data} is considered as a strong baseline for the non-NN based method to classify MTS datasets. 
In this section, we compare the performance of 
one neural network model which embeds the CSA module and the 1-Nearest-Neighbor classifier (with {\em DTW$_d$}) to show that the neural network based models with CSA generally outperform the Nearest-Neighbor approaches.
Without loss of generality and to save space, we use {\em MLSTM-FCN-CSA} as a representative for the neural network models with the CSA module. Note that the performance of the other NN-CSA models can be obtained from Table~\ref{tb:mts_details}.

Table~\ref{tb:dtw_fcnmlstm} shows the classification accuracy values of the two models. 
As shown in the table, 
the {\em MLSTM-FCN-CSA} performs better than the Nearest-Neighbor method ({\em DTW$_d$}) on most datasets (18/28) and has better averaged accuracy. Please be aware that optimizing the training of an NN-based model is not the focus of this paper. The results in this section are from a relatively small number of training epochs (400). The effect of the training epochs is analyzed in Section~\ref{sec:exp_parameter}.

\subsection{Case Study: Visual Examples of Class-Specific Features}
We conduct a case study to present the class-specific features using an example. It is challenging to generate user friendly feature examples from deep learning models. Please note that 
optimally visualizing deep learning features is not the focus of this work. 

We present two feature matrices ($P^{L}$ and $P^{O}$), before and after the CSA module, to show how the CSA module changes the regular features to class-specific features in one training batch (with batch size $B$). $P^{L}$ and $P^{O}$ are post-processed from $\mathcal{L}$ and $\mathcal{O}^{CSA}$ in Fig.~\ref{fig:csa_details}. 
%
 $P^{L}$ is obtained by conducting an  average pooling over the $T$ dimension of 
 $\mathcal{L} \in R^{B\times T \times F}$. 
 Thus, its size is $B\times F$. 
 Similarly, $P^{O}$ is generated from the average pooling over the $T$ dimension on $\mathcal{Q_{CSA}}$, 
 and has size 
$B\times C \times F$. 
For each training instance, 
$P^{O}$ has $C$ feature sequences whose length is $F$, while $P^{L}$ has only one feature sequence with length $F$.

\begin{figure}[htb]
\centering
\includegraphics[width=1\linewidth,height=0.6\linewidth]{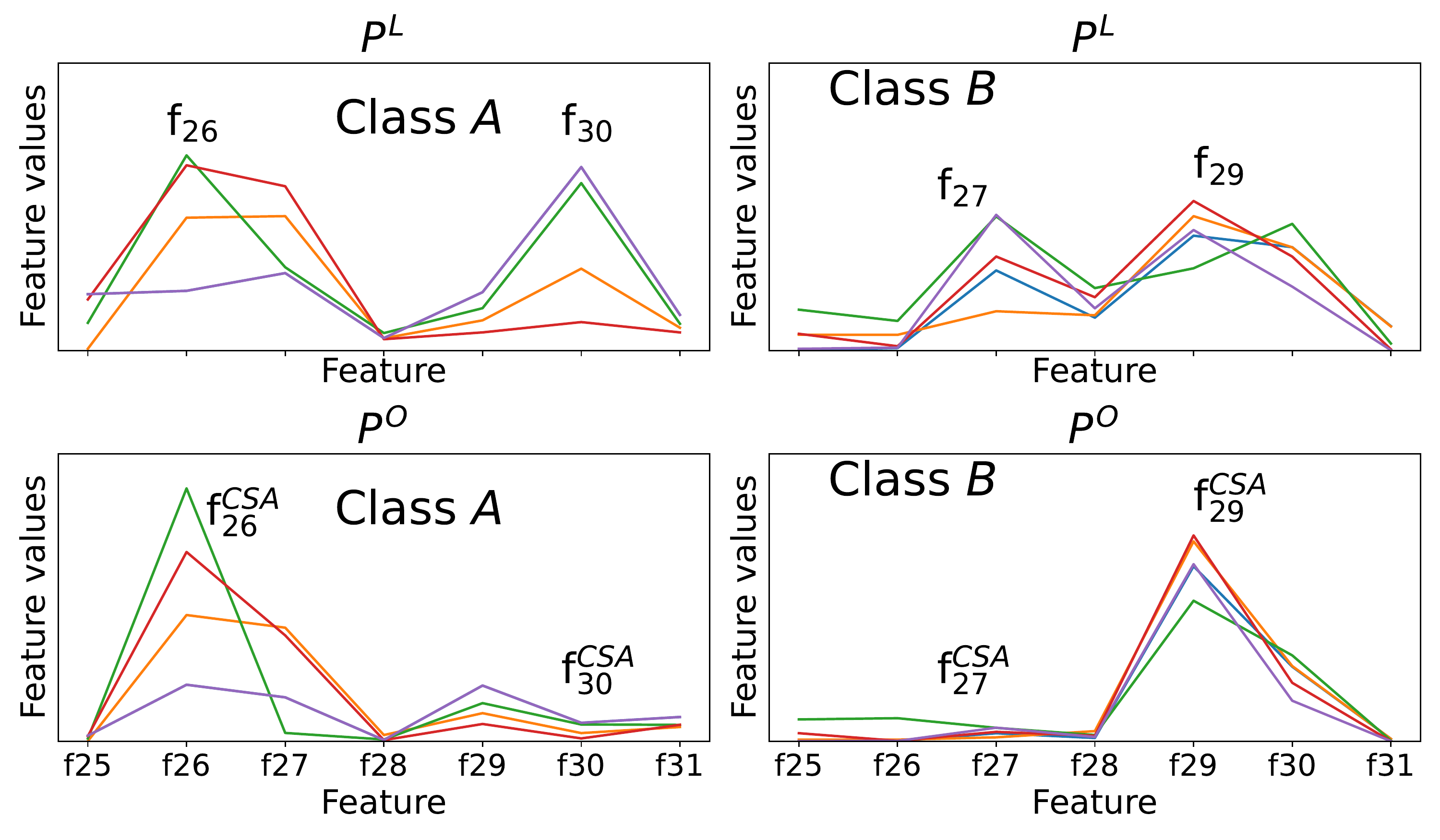}\\
{\small (a) Feature comparisons for class A and B}\\
\includegraphics[width=1\linewidth,height=0.6\linewidth]{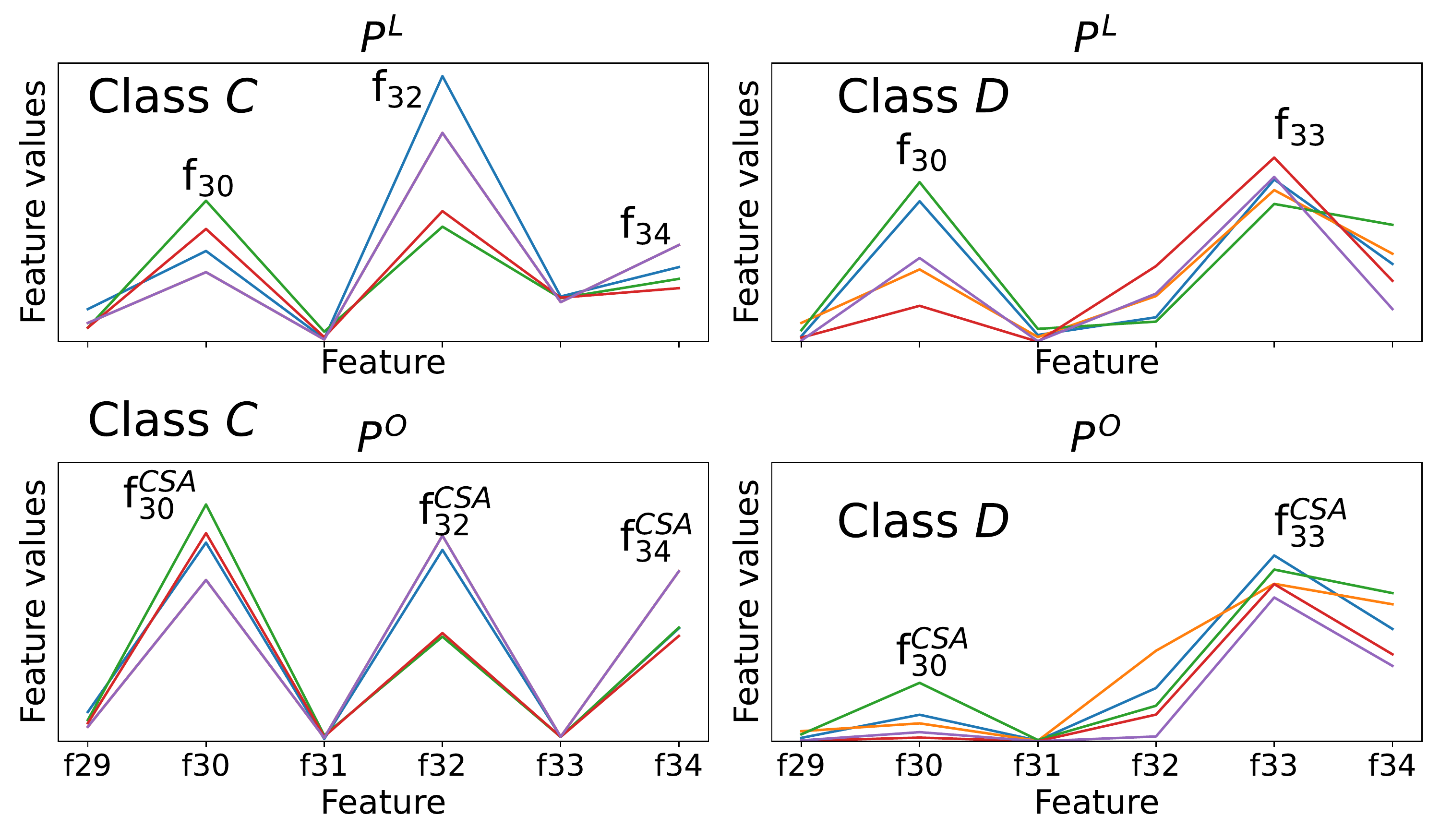}\\
{\small (b) Feature comparisons for class C and D}\\
\caption{Feature comparisons (5 random instances are drawn in different colors; first row plots regular features, second row plots class-specific features.)}
\label{fig:visual_exp_AB}
\end{figure}

For a specific class $c$, we compare the differences between the $c$th feature sequences (for $B$ instances) in $P^{O}$ with the feature sequences (for the same $B$ instances) in $P^{L}$.
Fig.~\ref{fig:visual_exp_AB} presents the comparison between two pairs of example classes. 
To improve the readability, we randomly pick 5 instances (in one batch) from classes $A$ and $B$, and only show a subset of features ($f_{25}$ to $f_{31}$).

We can see from the plotted features that the CSA module can emphasize some features (with higher values)    and weaken other features (with lower values). E.g., In Fig.~\ref{fig:visual_exp_AB}(a), feature $f_{30}$  is weakened (lower values) by the CSA module for class A and feature $f_{29}$ is emphasized by the CSA module for class B. 
The CSA module may change the importance of features. For example,
in Fig.~\ref{fig:visual_exp_AB}(b), $f_{30}$ was less important (smaller) than $f_{32}$ in the original feature space. However, the CSA module 
makes the $f_{30}^{CSA}$ more important (with higher value) than $f_{32}^{CSA}$.

\begin{table*}[htb]
\centering
\small
\begin{tabular}{l||| c | c || c ||| c | c || c ||| c | c || c}
\toprule
& \multicolumn{3}{c|||}{{\text{{\em FCN trained with 400 epochs}}}}
 & \multicolumn{3}{c|||}{{\text{{\em FCN trained with 800 epochs}}}}
 & \multicolumn{3}{c}{{\text{{\em FCN trained with 1200 epochs}}}}
 \\\cline{2-10}
 Datasets  & {\text{{\em w/o CSA}}} & {\text{{\em w CSA}}} & {\text{{\em AI}}} & {\text{{\em w/o CSA}}} & {\text{{\em w CSA}}} & {\text{{\em AI}}} & {\text{{\em w/o CSA}}} & {\text{{\em w CSA}}} & {\text{{\em AI}}} \\
\midrule
ArtWordRec        &   0.980  &  0.982  &  \bf{0.204}   &  0.980  &  0.986  &  \bf{0.612}   &  0.980  &  0.986  &  \bf{0.612}    \\
BasicMotions      &   0.968  &  0.966  &  -0.207  &  0.956  &  0.974  &  \bf{1.883}   &  0.968  &  0.980  &  \bf{1.240}    \\
CharTraj          &   0.990  &  0.990  &  0.000   &  0.990  &  0.990  &  0.000   &  0.990  &  0.990  &  0.000    \\
Cricket           &   0.910  &  0.910  &  0.000   &  0.900  &  0.928  &  \bf{3.111}   &  0.912  &  0.928  &  \bf{1.754}    \\
DuckDuckGeese     &   0.740  &  0.766  &  \bf{3.514}   &  0.750  &  0.746  &  -0.533  &  0.750  &  0.750  &  0.000    \\
EigenWorms        &   0.544  &  0.552  &  \bf{1.471}   &  0.566  &  0.586  &  \bf{3.534}   &  0.580  &  0.590  &  \bf{1.724}    \\
Epilepsy          &   0.794  &  0.842  &  \bf{6.045}   &  0.828  &  0.832  &  \bf{0.483}   &  0.844  &  0.856  &  \bf{1.422}    \\
EthanolConc       &   0.624  &  0.658  &  \bf{5.449}   &  0.646  &  0.674  &  \bf{4.334}   &  0.660  &  0.686  &  \bf{3.939}    \\
FaceDetection     &   0.562  &  0.562  &  0.000   &  0.562  &  0.564  &  \bf{0.356}   &  0.566  &  0.568  &  \bf{0.353}    \\
FingerMovements   &   0.648  &  0.656  &  \bf{1.235}   &  0.640  &  0.646  &  \bf{0.938}   &  0.642  &  0.650  &  \bf{1.246}    \\
HandMovement      &   0.468  &  0.488  &  \bf{4.274}   &  0.472  &  0.474  &  \bf{0.424}   &  0.456  &  0.484  &  \bf{6.140}    \\
Handwriting       &   0.284  &  0.288  &  \bf{1.408}   &  0.290  &  0.302  &  \bf{4.138}   &  0.290  &  0.298  &  \bf{2.759}    \\
Heartbeat         &   0.812  &  0.818  &  \bf{0.739}   &  0.812  &  0.820  &  \bf{0.985}   &  0.814  &  0.818  &  \bf{0.491}    \\
InsectWingbeat    &   0.108  &  0.138  &  \bf{27.778}  &  0.104  &  0.130  &  \bf{25.000}  &  0.100  &  0.142  &  \bf{42.000}   \\
JapaneseVowels    &   0.882  &  0.890  &  \bf{0.907}   &  0.880  &  0.891  &  \bf{1.250}   &  0.886  &  0.894  &  \bf{0.903}    \\
LSST              &   0.444  &  0.456  &  \bf{2.703}   &  0.462  &  0.472  &  \bf{2.165}   &  0.476  &  0.488  &  \bf{2.521}    \\
Libras            &   0.902  &  0.906  &  \bf{0.443}   &  0.908  &  0.908  &  0.000   &  0.896  &  0.908  &  \bf{1.339}    \\
MotorImagery      &   0.640  &  0.648  &  \bf{1.250}   &  0.642  &  0.646  &  \bf{0.623}   &  0.658  &  0.680  &  \bf{3.343}    \\
NATOPS            &   0.886  &  0.898  &  \bf{1.354}   &  0.898  &  0.900  &  \bf{0.223}   &  0.900  &  0.904  &  \bf{0.444}    \\
PEMS-SF           &   0.930  &  0.946  &  \bf{1.720}   &  0.942  &  0.948  &  \bf{0.637}   &  0.946  &  0.950  &  \bf{0.423}    \\
PenDigits         &   0.984  &  0.982  &  -0.203  &  0.984  &  0.982  &  -0.203  &  0.982  &  0.980  &  -0.204   \\
Phoneme           &   0.088  &  0.092  &  \bf{4.545}   &  0.094  &  0.098  &  \bf{4.255}   &  0.100  &  0.092  &  -8.000   \\
RacketSports      &   0.748  &  0.774  &  \bf{3.476}   &  0.776  &  0.788  &  \bf{1.546}   &  0.770  &  0.788  &  \bf{2.338}    \\
SelfRegSCP1       &   0.872  &  0.874  &  \bf{0.229}   &  0.886  &  0.866  &  -2.257  &  0.870  &  0.874  &  \bf{0.460}    \\
SelfRegSCP2       &   0.568  &  0.586  &  \bf{3.169}   &  0.572  &  0.586  &  \bf{2.448}   &  0.574  &  0.582  &  \bf{1.394}    \\
SpokenArab        &   0.974  &  0.982  &  \bf{0.821}   &  0.976  &  0.984  &  \bf{0.820}   &  0.976  &  0.980  &  \bf{0.410}    \\
StandWalkJump     &   0.400  &  0.400  &  0.000   &  0.414  &  0.480  &  \bf{15.942}  &  0.440  &  0.522  &  \bf{18.636}   \\
UWaveGesture      &   0.624  &  0.618  &  -0.962  &  0.612  &  0.614  &  \bf{0.327}   &  0.622  &  0.612  &  -1.608   \\\hline
Averaged    &     &    &  \bf{2.549}   &    &    &  \bf{2.609}   &    &    &  \bf{3.074}    \\
\bottomrule
\end{tabular}
\vspace{0.05in}
\caption{
Comparison of {\em FCN} and {\em FCN-CSA} 
with different number of epochs}
\label{tb:fcn_csa_epochs}
\end{table*}

\subsection{
Study of the Effect of Parameters
}
\label{sec:exp_parameter}
In this section, we study the effectiveness of the proposed CSA with different training epochs. There are many other neural network parameters, such as the number of layers, and the number of neurons, which are used to define the model architecture. We do not 
study the architecture parameters since optimizing the network structure is an open question and is not the focus of this paper. With the same model architecture ({\em Model} and {\em Model-CSA}), we would like to examine whether the {\em AI} improves 
with the increase of the  
number of epochs. 
We use {\em FCN} in this section as a representative neural network model.

In addition to the initial number of epochs (400)
used in the training (to report all the previous results), we increase the epoch number to 800 and 1200. Table~\ref{tb:fcn_csa_epochs} presents the accuracy and {\em AI} results from {\em FCN} and {\em FCN-CSA} models. The results show that the performance improvements from the CSA module are consistent with all 3 epoch numbers. It means that the CSA module indeed learns 
novel and helpful features, which 
cannot be well learned by the base {\em FCN} model 
even with more training epochs.

For a given individual dataset, it is not guaranteed that the model has better performance with higher epoch numbers.  
The CSA improvements (accuracy and {\em AI}) are not monotonically increasing with the increase of epoch numbers. 
This is caused by the slight instability of deep-learning models after the model converges.
However, the overall (averaged) results over all MTS datasets (last row in Table~\ref{tb:fcn_csa_epochs}) with higher epoch number get better than the results with lower epoch.

\begin{figure}[htb]
\centering
\includegraphics[width=0.9\linewidth,height=0.33\linewidth]{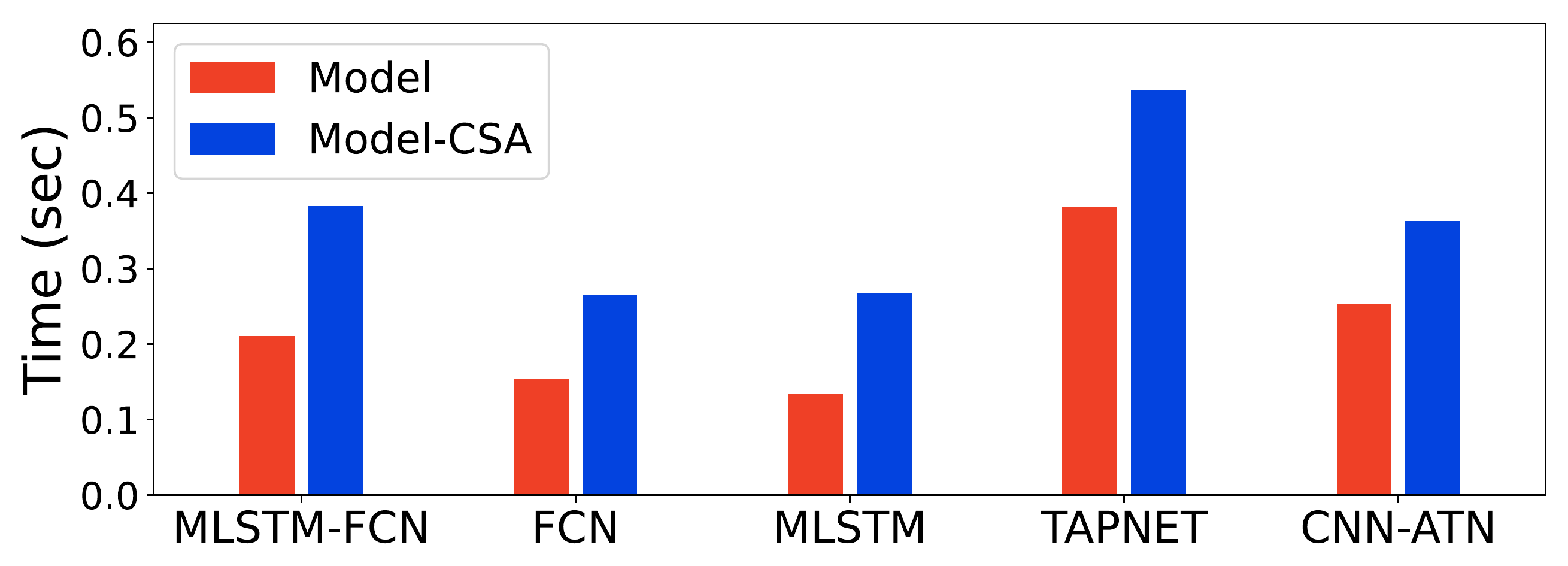}\\ (a) Training time per iteration
\includegraphics[width=0.9\linewidth,height=0.33\linewidth]{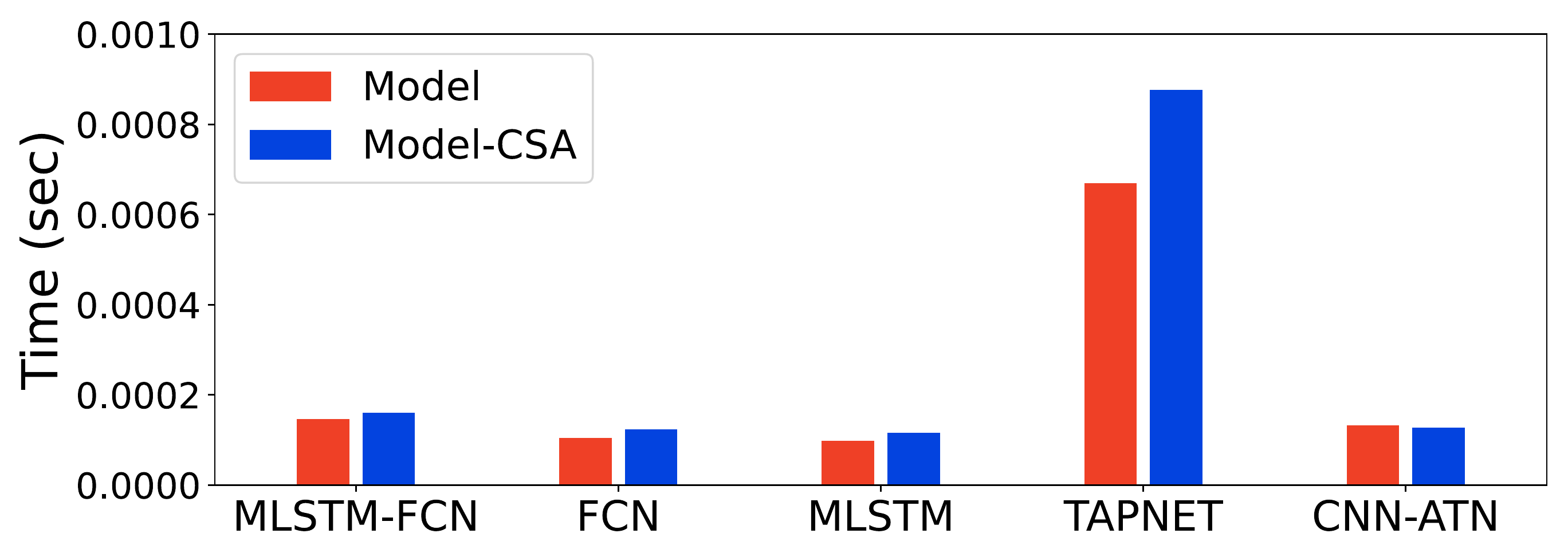}\\(b) Testing time per instance
\caption{Running time (averaged on all the datasets)}
\label{fig:time}
\end{figure}
\subsection{Efficiency Analysis}
The above results show that class-specific attention generally has  positive impact on classification accuracy. In this section, we examine the  efficiency of the proposed  CSA module.
 
 Fig.~\ref{fig:time} (a) reports the training time 
 for the baseline models (i.e.,  {\em Model}) 
 and {\em Model-CSA} per iteration.
%
%
As we would expect, CSA requires additional time to calculate the tensors and learning parameters in the CSA architecture. {\em Model-CSA} uses up to 0.2 seconds more than {\em Model} to train the models in one iteration.
%
%
While class-specific attention has an impact on training times, 
there are no significant differences between the testing times of baseline models with models extended with 
CSA. Fig.~\ref{fig:time} (b) shows
{\em Model-CSA} 
uses almost the same time as 
{\em Model} to make predictions for one instance for all models except  {\em TapNet}. {\em TapNet} has a bigger difference because it has more features (three times) than other models to learn in the CSA module.
This confirms that class-specific attention can be used effectively in practice for accurate time series classification.

\section{Conclusions and Future Works}
\label{sec:con}
This paper presents a class-specific attention (CSA) module to improve the performance of neural network models for time series classification. 
The proposed module can be embedded to  neural network models for time series classification (including those that leverage other forms of attention) to automatically capture significant features to differentiate 
instances of one class from the instances of the other classes.
%
CSA identifies class-specific features leveraging training labels, while avoiding the need to access label information during testing phase.
To the best of our knowledge, this is the first attention design that leverages the class label information in the hidden layers to generate class-specific features.
%
Experiments on 40 real datasets have shown that CSA generally boosts accuracy -- in the experiments, we have seen performance improvements up to 42\%. 
Statistical analysis shows that the CSA module improves a base neural network model on 67\% of MTS tests and 80\% of the UTS tests, is significantly better than a base base on 11\% of MTS and 13\% of UTS tests. 

To make the tests more extensive, we can apply the models on more time series datasets (e.g., all the UTS datasets in the UCR repository). As a further step, the same design principle of CSA module can be utilized to improve the classification performance of other types of data such as images.

\section*{Acknowledgment}
This research has been supported by NSF awards 163330, 1914635, and 1757207. 

\bibliographystyle{IEEEtran}
\bibliography{reference}

\begin{thebibliography}{10}
\providecommand{\url}[1]{#1}
\csname url@samestyle\endcsname
\providecommand{\newblock}{\relax}
\providecommand{\bibinfo}[2]{#2}
\providecommand{\BIBentrySTDinterwordspacing}{\spaceskip=0pt\relax}
\providecommand{\BIBentryALTinterwordstretchfactor}{4}
\providecommand{\BIBentryALTinterwordspacing}{\spaceskip=\fontdimen2\font plus
\BIBentryALTinterwordstretchfactor\fontdimen3\font minus
  \fontdimen4\font\relax}
\providecommand{\BIBforeignlanguage}[2]{{%
\expandafter\ifx\csname l@#1\endcsname\relax
\typeout{** WARNING: IEEEtran.bst: No hyphenation pattern has been}%
\typeout{** loaded for the language `#1'. Using the pattern for}%
\typeout{** the default language instead.}%
\else
\language=\csname l@#1\endcsname
\fi
#2}}
\providecommand{\BIBdecl}{\relax}
\BIBdecl

\bibitem{ref:lstm_neco1997}
\BIBentryALTinterwordspacing
S.~Hochreiter and J.~Schmidhuber, ``Long short-term memory,'' \emph{Neural
  Computation}, vol.~9, no.~8, pp. 1735--1780, 1997. [Online]. Available:
  \url{https://doi.org/10.1162/neco.1997.9.8.1735}
\BIBentrySTDinterwordspacing

\bibitem{ref:dcnn_ijcai2015}
\BIBentryALTinterwordspacing
J.~Yang, M.~N. Nguyen, P.~P. San, X.~Li, and S.~Krishnaswamy, ``Deep
  convolutional neural networks on multichannel time series for human activity
  recognition,'' in \emph{{IJCAI}}, 2015, pp. 3995--4001. [Online]. Available:
  \url{http://ijcai.org/Abstract/15/561}
\BIBentrySTDinterwordspacing

\bibitem{ref:fcn_ijcnn2017}
\BIBentryALTinterwordspacing
Z.~Wang, W.~Yan, and T.~Oates, ``Time series classification from scratch with
  deep neural networks: {A} strong baseline,'' in \emph{International Joint
  Conference on Neural Networks ({IJCNN})}, 2017, pp. 1578--1585. [Online].
  Available: \url{https://doi.org/10.1109/IJCNN.2017.7966039}
\BIBentrySTDinterwordspacing

\bibitem{ref:cross_attn_ijcai2020}
\BIBentryALTinterwordspacing
Y.~Hao and H.~Cao, ``A new attention mechanism to classify multivariate time
  series,'' in \emph{{IJCAI}}, 2020, pp. 1999--2005. [Online]. Available:
  \url{https://doi.org/10.24963/ijcai.2020/277}
\BIBentrySTDinterwordspacing

\bibitem{ref:lstm_fcn_acc2018}
\BIBentryALTinterwordspacing
F.~Karim, S.~Majumdar, H.~Darabi, and S.~Chen, ``{LSTM} fully convolutional
  networks for time series classification,'' \emph{{IEEE} Access}, vol.~6, pp.
  1662--1669, 2018. [Online]. Available:
  \url{https://doi.org/10.1109/ACCESS.2017.2779939}
\BIBentrySTDinterwordspacing

\bibitem{ref:mlstm_fcn_nn2019}
\BIBentryALTinterwordspacing
F.~Karim, S.~Majumdar, H.~Darabi, and S.~Harford, ``Multivariate lstm-fcns for
  time series classification,'' \emph{Neural Networks}, vol. 116, pp. 237--245,
  2019. [Online]. Available: \url{https://doi.org/10.1016/j.neunet.2019.04.014}
\BIBentrySTDinterwordspacing

\bibitem{ref:mccnn_fcs2016}
\BIBentryALTinterwordspacing
Y.~Zheng, Q.~Liu, E.~Chen, Y.~Ge, and J.~L. Zhao, ``Exploiting multi-channels
  deep convolutional neural networks for multivariate time series
  classification,'' \emph{Frontiers Comput. Sci.}, vol.~10, no.~1, pp. 96--112,
  2016. [Online]. Available: \url{https://doi.org/10.1007/s11704-015-4478-2}
\BIBentrySTDinterwordspacing

\bibitem{ref:psv_tkde2019}
Y.~Hao, H.~Cao, A.~Mueen, and S.~Brahma, ``Identify significant
  phenomenon-specific variables for multivariate time series,'' \emph{{IEEE}
  Trans. Knowl. Data Eng.}, vol.~31, pp. 1--13, 2019.

\bibitem{ref:tapnet_aaai2020}
\BIBentryALTinterwordspacing
X.~Zhang, Y.~Gao, J.~Lin, and C.~Lu, ``Tapnet: Multivariate time series
  classification with attentional prototypical network,'' in \emph{The
  Thirty-Fourth Conference on Artificial Intelligence, {AAAI} 2020, February
  7-12, 2020}, 2020, pp. 6845--6852. [Online]. Available:
  \url{https://aaai.org/ojs/index.php/AAAI/article/view/6165}
\BIBentrySTDinterwordspacing

\bibitem{ref:first_attn_iclr2015}
\BIBentryALTinterwordspacing
D.~Bahdanau, K.~Cho, and Y.~Bengio, ``Neural machine translation by jointly
  learning to align and translate,'' in \emph{{ICLR}}, 2015. [Online].
  Available: \url{http://arxiv.org/abs/1409.0473}
\BIBentrySTDinterwordspacing

\bibitem{ref:google_all_atte_nips2017}
\BIBentryALTinterwordspacing
A.~Vaswani, N.~Shazeer, N.~Parmar, J.~Uszkoreit, L.~Jones, A.~N. Gomez,
  L.~Kaiser, and I.~Polosukhin, ``Attention is all you need,'' in \emph{Annual
  Conference on Neural Information Processing Systems 2017}, 2017, pp.
  5998--6008. [Online]. Available:
  \url{http://papers.nips.cc/paper/7181-attention-is-all-you-need}
\BIBentrySTDinterwordspacing

\bibitem{ref:drnn_arxiv2014}
\BIBentryALTinterwordspacing
R.~Pascanu, {\c{C}}.~G{\"{u}}l{\c{c}}ehre, K.~Cho, and Y.~Bengio, ``How to
  construct deep recurrent neural networks,'' in \emph{2nd International
  Conference on Learning Representations ({ICLR})}, 2014. [Online]. Available:
  \url{http://arxiv.org/abs/1312.6026}
\BIBentrySTDinterwordspacing

\bibitem{ref:dual_attn_rnn_ijcai2017}
\BIBentryALTinterwordspacing
Y.~Qin, D.~Song, H.~Chen, W.~Cheng, G.~Jiang, and G.~W. Cottrell, ``A
  dual-stage attention-based recurrent neural network for time series
  prediction,'' in \emph{{IJCAI}}, 2017, pp. 2627--2633. [Online]. Available:
  \url{https://doi.org/10.24963/ijcai.2017/366}
\BIBentrySTDinterwordspacing

\bibitem{ref:gru_2014}
\BIBentryALTinterwordspacing
K.~Cho, B.~van Merrienboer, D.~Bahdanau, and Y.~Bengio, ``On the properties of
  neural machine translation: Encoder-decoder approaches,'' \emph{CoRR}, vol.
  abs/1409.1259, 2014. [Online]. Available:
  \url{http://arxiv.org/abs/1409.1259}
\BIBentrySTDinterwordspacing

\bibitem{ref:attn_voice_2017}
\BIBentryALTinterwordspacing
C.~Shan, J.~Zhang, Y.~Wang, and L.~Xie, ``Attention-based end-to-end speech
  recognition in mandarin,'' \emph{CoRR}, vol. abs/1707.07167, 2017. [Online].
  Available: \url{http://arxiv.org/abs/1707.07167}
\BIBentrySTDinterwordspacing

\bibitem{ref:cnn_class_specific}
Y.~Hao, H.~Cao, and E.~Draayer, ``Cnn approaches to classify multivariate time
  series using class-specific features,'' in \emph{2020 IEEE International
  Conference on Smart Data Services (SMDS)}, 2020, pp. 1--8.

\bibitem{ref:uea_data}
A.~Bagnall, H.~A. Dau, J.~Lines, M.~Flynn, J.~Large, A.~Bostrom, P.~Southam,
  and E.~Keogh, ``The {UEA} multivariate time series classification archive,
  2018,'' 2018.

\bibitem{ref:ucr_data}
Y.~Chen, E.~Keogh, B.~Hu, N.~Begum, A.~Bagnall, A.~Mueen, and G.~Batista, ``The
  ucr time series classification archive,'' July 2015,
  \url{www.cs.ucr.edu/~eamonn/time\_series\_data/}.

\bibitem{ref:null_hypothesis}
C.~Pernet, ``Null hypothesis significance testing: a short tutorial,''
  \emph{F1000Research}, vol.~4, 08 2015.

\bibitem{ref:pearson_chisquare}
\BIBentryALTinterwordspacing
K.~Pearson, ``X. on the criterion that a given system of deviations from the
  probable in the case of a correlated system of variables is such that it can
  be reasonably supposed to have arisen from random sampling,'' \emph{The
  London, Edinburgh, and Dublin Philosophical Magazine and Journal of Science},
  vol.~50, no. 302, pp. 157--175, Jul. 1900. [Online]. Available:
  \url{https://doi.org/10.1080/14786440009463897}
\BIBentrySTDinterwordspacing

\bibitem{ref:knn}
\BIBentryALTinterwordspacing
A.~Mucherino, P.~J. Papajorgji, and P.~M. Pardalos, \emph{k-Nearest Neighbor
  Classification}.\hskip 1em plus 0.5em minus 0.4em\relax New York, NY:
  Springer New York, 2009, pp. 83--106. [Online]. Available:
  \url{https://doi.org/10.1007/978-0-387-88615-2\_4}
\BIBentrySTDinterwordspacing

\bibitem{ref:dtw}
T.~Giorgino, ``Computing and visualizing dynamic time warping alignments in
  {R}: The {dtw} package,'' \emph{Journal of Statistical Software}, vol.~31,
  no.~7, pp. 1--24, 2009.

\bibitem{DBLP:journals/pvldb/DingTSWK08}
\BIBentryALTinterwordspacing
H.~Ding, G.~Trajcevski, P.~Scheuermann, X.~Wang, and E.~J. Keogh, ``Querying
  and mining of time series data: experimental comparison of representations
  and distance measures,'' \emph{Proc. {VLDB} Endow.}, vol.~1, no.~2, pp.
  1542--1552, 2008. [Online]. Available:
  \url{http://www.vldb.org/pvldb/vol1/1454226.pdf}
\BIBentrySTDinterwordspacing

\end{thebibliography}

\end{document}